\documentclass{article}

\PassOptionsToPackage{usenames,dvipsnames}{xcolor}

\PassOptionsToPackage{numbers, compress}{natbib}

    \usepackage[final]{neurips_2024}

\PassOptionsToPackage{frozencache,cachedir=.}{minted}

\PassOptionsToPackage{final}{showkeys}

\usepackage{garyw}
\usepackage{titletoc}
\usepackage{tablefootnote}
\usepackage{wrapfig}
\usepackage{nicefrac}
\hypersetup{
colorlinks=true,
allcolors=RoyalBlue
}

\newcommand{\chemflow}{\textbf{\textit{ChemFlow}}}

\definecolor{spv_green}{HTML}{66CC00}
\definecolor{unsup_yellow}{HTML}{CCCC00}

\DeclareMathOperator{\logp}{logP}
\DeclareMathOperator{\sa}{SA}

\title{Navigating Chemical Space with Latent Flows}

\author{Guanghao Wei\textsuperscript{*$\ddagger$}\\
Cornell University\\
\texttt{\small gw338@cornell.edu}
\And
Yining Huang\textsuperscript{*}\\
Harvard University\\
\texttt{\small yininghuang@hms.harvard.edu}
\AND
Chenru Duan\\
Deep Principle, Inc.\\
\texttt{\small duanchenru@gmail.com}
\And
Yue Song\textsuperscript{$\dagger$}\\
California Institute of Technology\\
\texttt{\small yuesong@caltech.edu}
\And
Yuanqi Du\textsuperscript{$\dagger$}\\
Cornell University\\
\texttt{\small yd392@cornell.edu}
}

\date{} %

\begin{document}

\maketitle

\customfootnotetext{*}{ Equal Contribution.}
\customfootnotetext{$\dagger$}{ Equal Supervision.}
\customfootnotetext{$\ddagger$}{ This work was completed while the author was at Cornell University.}

\begin{abstract}
Recent progress of deep generative models in the vision and language domain has stimulated significant interest in more structured data generation such as molecules. However, beyond generating new random molecules, efficient exploration and a comprehensive understanding of the vast chemical space are of great importance to molecular science and applications in drug design and materials discovery.
In this paper, we propose a new framework, ChemFlow, to traverse chemical space through navigating the latent space learned by molecule generative models through flows. We introduce a dynamical system perspective that formulates the problem as learning a vector field that transports the mass of the molecular distribution to the region with desired molecular properties or structure diversity. 
Under this framework, we unify previous approaches on molecule latent space traversal and optimization and propose alternative competing methods incorporating different physical priors. 
We validate the efficacy of ChemFlow on molecule manipulation and single- and multi-objective molecule optimization tasks under both supervised and unsupervised molecular discovery settings.
Codes and demos are publicly available on GitHub at 
\href{https://github.com/garywei944/ChemFlow}{https://github.com/garywei944/ChemFlow}.
\end{abstract}

\section{Introduction}

Designing new functional molecules has been a long-standing challenge in molecular discovery which concerns a wide range of applications in drug design and materials discovery~\cite{sanchez2018inverse,vamathevan2019applications}. With the increasing interest in applying deep learning models in scientific problems~\cite{wang2023scientific,zhang2023artificial}, molecular design has attracted considerable attention given its massively available data and accessible evaluations. Among the developed methods, two paradigms emerge: one paradigm searches for new molecules based on combinatorial optimization approaches respecting the discrete nature of molecules; the other paradigm builds upon the success of deep generative models in approximating the molecular distribution with a given dataset and then generating new molecules from the learned models~\cite{du2022molgensurvey}. Both of the approaches have demonstrated promising results in small molecule, protein, and materials design~\cite{watson2023novo,ingraham2023illuminating,loeffler2023reinvent4,zeni2023mattergen}.
Despite the promise, the chemical space is tremendously large with the number of drug-like small molecule compounds estimated to be from $10^{23}$ to $10^{60}$~\cite{bohacek1996art,lipinski2012experimental}. This necessitates either more efficient searching methods or better understanding about the structure of the chemical space. Following the progress made in studying the latent structure of deep generative models
(\textit{e.g.} generative adversarial networks~(GANs)~\cite{goodfellow2016deep}, variational autoencoders~(VAEs)~\cite{kingma2013auto}, and denoising diffusion models~\cite{ho2020denoising})
in computer vision~\cite{jahanian2019steerability,Burgess2018UnderstandingDI,harkonen2020ganspace,kwon2022diffusion}, decent efforts have recently been made in understanding the learned latent space of molecule generative models. 

Initially, disentangled representation learning becomes a popular paradigm to enforce a structured and interpretable representation~\cite{du2022interpretable}.
Specifically, each latent dimension is expected to learn a disentangled factor of variation, and tweak the latent vector along the dimension could lead to generating new samples with changes only in one molecular property.
However, even if imposing such constraints in the training of molecule generative models, the models still struggle to learn meaningful disentangled factors in the early attempts~\cite{du2022small}.
In addition to constraining the model training procedure, exploring the structure of pre-trained molecule generative models is more efficient.
The main approach developed is to utilize optimization approaches to discover the region in the latent space with the desired molecular property. It often trains a proxy function to map from the latent vector to the property, providing access to gradients for gradient-based optimization~\cite{liu2018constrained,griffiths2020constrained,eckmann2022limo}. The third line of work also builds upon pre-trained models as well. It leverages one interesting finding such that the learned latent space of molecule generative models is linearly separable~\cite{gomez2018automatic}, which is also widely studied and used as a priori in computer vision~\cite{shen2020interfacegan,shen2021closed}. ChemSpace~\cite{Du2023ChemSpacEIA} develops a highly efficient approach to use linear classifiers to identify the separation boundary and considers the normal direction to the boundary as the direction of control. Nevertheless, the linear separability assumption may be too strong.
It is worth noting that the first line of work does not require labels and can be trained in an unsupervised manner (referred to as unsupervised discovery), while both the second and third lines of work require access to labels to train/identify a guidance model/direction (referred to as supervised discovery).

In this paper, we propose a new framework, \chemflow{}, based on flows in a dynamical system to efficiently explore the latent structure of molecule generative models.
Specifically, we unify previous approaches (gradient-based optimization, linear latent traversal, and disentangled traversal) under the realm of flows that transforms data density along time via a vector field.
In contrast to previous linear models, our framework is flexible to learn nonlinear transformations inspired by partial differential equations~(PDEs) governing real-world physical systems such as heat and wave equations.
We then analyze how different dynamics may bring special properties to solve different tasks.
Our framework can also generalize both supervised and unsupervised settings under the same umbrella.
Particularly in the under-studied unsupervised setting, we demonstrate a structure diversity potential can be incorporated to find trajectories that maximize the structure change of the molecules (which in turn leads to property change).
We conduct extensive experiments with physicochemical, drug-related properties, and protein-ligand binding affinities on both molecule manipulation and (single- and multi-objective) molecule optimization experiments.
The experiment results demonstrate the generality of the proposed framework and the effectiveness of alternative methods under this framework to achieve better or comparable results with existing approaches.

\section{Background}

\subsection{Navigating Latent Space of Molecules}
\label{sec:related_traversal}

The latent space $\gZ$ of molecule generative models is often learned through an encoder function $f_\theta(\cdot)$ and a decoder function $g_\psi(\cdot)$ such that the encoder maps the input molecular structures $\vx \in \mathcal{X}$ into an (often) low-dimensional and continuous space (\textit{i.e.} latent space) while the decoder maps the latent vectors $\vz \in \mathcal{Z}$ back to molecular structures $x'$. Note that this encoder-decoder architecture is general and can be realized by popular generative models such as VAEs, flow-based models, GANs, and diffusion models~\cite{jin2018junction,madhawa2019graphnvp,cao2018molgan,vignac2023digress,you2024latent}.
For simplicity, we focus on VAE-based methods in this paper. To traverse the learned latent space of molecule generative models, two approaches have been proposed: gradient-based optimization and latent traversal. 

The gradient-based optimization methods first learn a proxy function $h(\cdot)$ parameterized by a neural network that provides the direction to traverse~\cite{zang2020moflow}. This can be formulated as a gradient flow following the direction of steepest descent of the potential energy function $h(\cdot)$ and discretized, as follows:
\begin{equation}
\begin{aligned}
\rd\vz_t &= -\nabla_z h(\vz_t) \rd t \\
\vz_t &= \vz_{t-1}-\nabla_z h(\vz_{t-1})\rd t
\end{aligned}
\end{equation}
where we take a dynamic system perspective on the evolution of latent samples. The latent traversal approaches leverage the observation of linear separability in the learned latent space of molecule generative models~\cite{gomez2018automatic}. Since the direction is assumed to be linear, it can be found easily. ChemSpace~\cite{Du2023ChemSpacEIA} learns a linear classifier that defines the separation boundary of the molecular properties. Then the normal direction of the boundary provides a linear direction $\vn\in\gZ$ for traversing the latent space:
\begin{equation}
\vz_t = \vz_0 + \vn t
\end{equation}
We notice that the above gradient flow and linear traversal can be analyzed and designed from a dynamical system perspective: linear traversal can be considered as a special case of wave functions, \textit{i.e.}, we have $\nicefrac{\partial^2 \mathbf{z}_t}{\partial^2 \mathbf{z}_{t-1}} = \nicefrac{\partial^2 \mathbf{z}_t}{\partial^2 t}=0$ satisfied by wave functions. This connection inspires us to consider designing more dynamical traversal approaches in the latent space.

\subsection{Wasserstein Gradient Flows}
Gradient flows define the curve $\vx(t)\in \sR^n$ that evolves in the negative gradient direction of a function $\gF:\sR^n\rightarrow\sR$. The time evolution of the gradient flow is given by the ODE $\vx'(t)=-\nabla \gF(\vx(t))$. Wasserstein gradient flows describe a special type of gradient flow where $\gF$ is set to be the Wasserstein distance. For example, as introduced in~\citet{benamou2000computational}, the commonly used $L^2$ Wasserstein metric induces a dynamic formulation of optimal transport:
\begin{equation}
\label{eqn:gflow}
\begin{gathered}
    W_2(\mu,\nu)^2 = \min_{v, \rho} \Big\{\int\int \frac{1}{2}\rho(t,\vx)|v(t,\vx)|^2 \mathop{dt}\mathop{d\vx}:\partial_t\rho(t,\vx) = -\nabla\cdot(v(t,\vx)\rho(t,\vx)) \Big\}
\end{gathered}
\end{equation}
where $\mu,\nu$ are two probability measures at the source and target distributions, respectively. Interestingly, if we take the gradient of a potential energy $\nabla \phi$ as the velocity field applied to a distribution, the time evolution of $\nabla \phi$ can be seen to minimize the Wasserstein distance and thus follow optimal transport. In \cref{sec:app_wasserstein}, we give detailed derivations of how the vector fields minimize the $L^2$ Wasserstein distance and discuss alternative PDEs of the density evolution recovered by Wasserstein gradient flow (\textit{e.g.} Wasserstein gradient flow over the entropy functional recovers heat equation) following the seminal JKO scheme~\cite{Jordan1996THEVF}. 

\begin{figure*}[t]
\begin{center}
\includegraphics[width=0.95\linewidth]{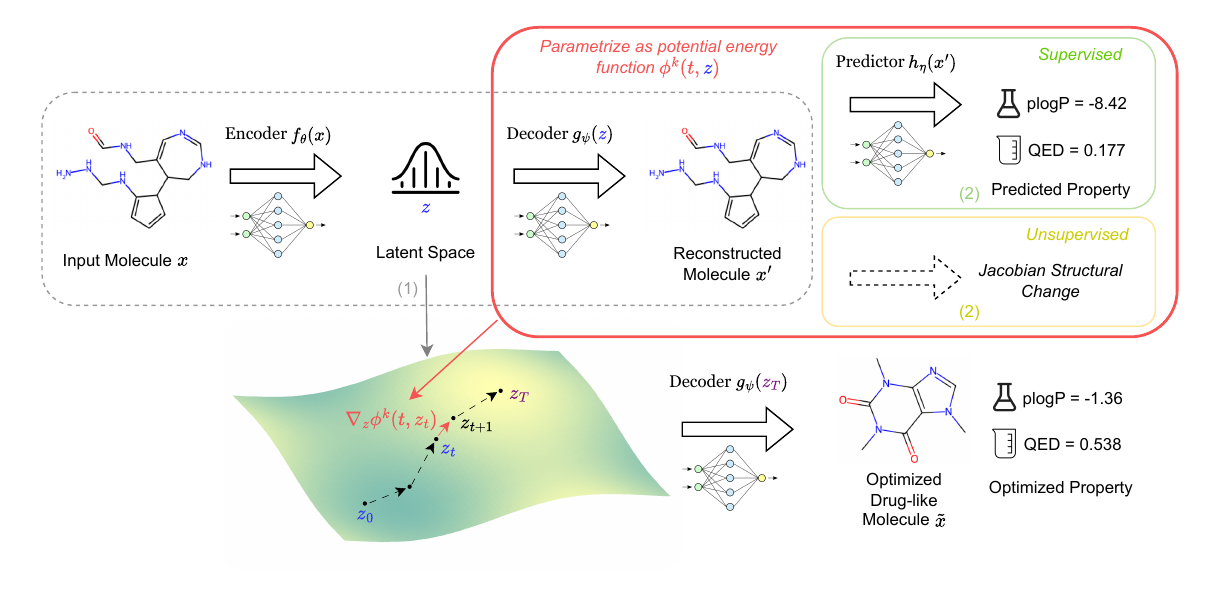}
\end{center}
\caption{
\textbf{\chemflow{}} framework: (1) a pre-trained encoder $f_\theta(\cdot)$ and decoder $g_\psi(\cdot)$ that maps between molecules $\vx$ and latent vectors $\vz$, (2) we use a property predictor $h_\eta(\cdot)$ (\textcolor{spv_green}{green box}) or a ``Jacobian control'' (\textcolor{unsup_yellow}{yellow box}) as the guidance to learn a vector field $\nabla_z\phi^k(t, \vz_t)$ that maximizes the change in certain molecular properties (e.g. plogP, QED) or molecular structures, (3) during the training process, we add additional dynamical regularization on the flow. The learned flows move the latent samples to change the structures and properties of the molecules smoothly.
(Better seen in color).
The flow chart illustrates a case where a molecule is manipulated into a drug like caffeine.
}
\label{fig:workflow}
\end{figure*}

\section{Methodology}
\label{sec:method}

We present \chemflow{} as a unified framework for latent traversals in chemical latent space as latent flows.
We parameterize a set of scalar-valued energy functions $\phi^k = \texttt{MLP}_{\theta^k}(t,\vz)\in\mathbb{R}$ and use the learned flow $\nabla_\vz \phi^k$ to traverse the latent samples. The traversal process can be described by the following equation in a Lagrangian way (particle trajectory):
\begin{equation}
\begin{split}
\vz_{t} = \vz_{t-1} + \nabla_\vz \phi^k(t-1, \vz_{t-1})
\end{split}
\end{equation}

Alternatively, as an Eulerian approach, we can write the time evolution of the density through a pushforward map:
\begin{equation}
    \rho_t = [\psi_t]_*\rho_{t-1}
\end{equation}
where $\psi_t$ defines the time-dependent flow that transforms the densities of latent samples through a probability path. The pushforward measure $[\psi_t]_*$ induces a change of variable formula for densities~\cite{rezende2015variational}:
\begin{equation}
    [\psi_t]_*\rho_{t-1}(\vz) = \rho_{t-1}(\psi_t^{-1}(\vz)) |\det \Big[\frac{\partial\psi_t^{-1}(\vz)}{\partial\vz}\Big]|
\end{equation}
In the following, we will introduce how $\nabla_\vz \phi^k$ is matched to some pre-defined velocities for generating different flows.

\subsection{Learning Different Latent Flows}
\label{sec:flows}

Given a pre-trained molecule generative model $g_\psi: \gZ \rightarrow \gX$ with prior distribution $p(\vz)$, we would like to model $K$ different semantically disentangled latent trajectories that correspond to different properties of the molecules, numbered by superscript $k$. 

\noindent\textbf{Hamilton-Jacobi Flows.} One desired property for the latent traversal comes from optimal transport theory such that the transport cost is minimized (i.e. shortest path). This property can be enforced by solving \Cref{eqn:gflow} by Karush–Kuhn–Tucker~(KKT) conditions, which will give the optimal solution — the Hamilton-Jacobi Equation~(HJE):
\begin{equation}\label{eq:hj}
    \frac{\partial}{\partial t} \phi^k(t,\vz) + \frac{1}{2} ||\nabla_\vz \phi^k(t,\vz)||^2 = 0
\end{equation}
where the velocity field is defined as the flow $\nabla \phi^k$. The HJE can also be interpreted as mass transportation in fluid dynamics, \textit{i.e.}, under the velocity field $\nabla \phi^k$, the fluid will evolve to the target distribution with an optimal transportation cost.

We achieve the HJE constraint by matching our flow fields and define the boundary condition as:
\begin{equation}
\begin{split}
\gL_r = \frac{1}{T}\sum^{T-1}_{t=0} \big(\frac{\partial}{\partial t} \phi^k(t,\vz) + \frac{1}{2} ||\nabla_\vz \phi^k(t,\vz)||^2 \big)^2,\  
\gL_\phi = \sum_{k=0}^{K-1}||\nabla_\vz \phi^k(0,\vz_0)||_2^2 \\
\end{split}
\end{equation}
where $T$ represents the total number of traversal steps, $\gL_r$ restricts the energy to obey our physical constraints, and $\gL_\phi$ restricts $\phi(t,\vz_t)$ to match the initial condition. Our latent traversal can be thus regarded as dynamic optimal transport between distributions of molecules with different properties. 

\noindent\textbf{Wave Flows.} Alternatively, we can pivot the optimal transport property to enforce additional physical and dynamical priors. For example, if we specify the flow to follow wave-like dynamics, we can use the second-order wave equation: 
\begin{equation}
\label{eq:wave}
\begin{split}
\frac{\partial^2}{\partial t^2} \phi^k(t, \vz) 
 - c^2 \nabla^2_\vz \phi^k(t, \vz) &= 0
\end{split}
\end{equation}
The above constraint empirically produces highly diverse and realistic trajectories. Our velocity matching objective and boundary condition then become:
\begin{equation}
\label{eq:lr_lphi}
\begin{split}
\gL_r = \frac{1}{T}\sum^{T-1}_{t=0} ||\frac{\partial^2}{\partial t^2} \phi^k(t, \vz) 
 - c^2 \nabla^2_\vz \phi^k(t, \vz) ||_2^2,\  
\gL_\phi = \sum_{k=0}{K-1}||\nabla_\vz \phi^k(0,\vz_0)||_2^2
\end{split}
\end{equation}
where $\gL_r$ and $\gL_\phi$ restrict the physical constraints and the initial condition, respectively. Note that $\phi^k\equiv0$ is a trivial optimal solution for the above two objectives regarding that both $\gL_r$ and $\gL_\phi$ are non-negative. To prevent the parameterized $\phi^k$ from converging to such a trivial solution, we introduce more guidance to the loss function in \Cref{sec:guidance} separately. 

\noindent\textbf{Alternative Flows.} Besides HJE and wave equations, our framework is also general to include other commonly used PDEs that allow for different dynamics along the flow, such as Fokker Planck equation and heat equation. In the experimental section, we will explore the effectiveness of each latent flow in different supervision settings.

\subsection{Supervised \& Unsupervised Guidance}
\label{sec:guidance}

\noindent\textbf{Supervised Semantic Guidance.} When an explicit semantic potential energy function or labeled data for the semantic of interest is available, we can use the provided semantic potential to guide the learning of the flow. Firstly, we train a surrogate model $h_\eta: \mathcal{X} \rightarrow \mathbb{R}$ (parameterized by a deep neural network) to predict the corresponding molecular property. Then we use the trained surrogate model as guidance to learn flows that drive the increase of the property for the trajectory of the generated molecules.
\begin{equation}
    \begin{aligned}
    d = \inner{-\nabla_{\vz} h_\eta(g_\psi(\vz_{t}))}{\nabla_\vz \phi^k(t, \vz_{t})},\ 
    \gL_{\gP} = -\sign(d)\norm{d}_2^2
\end{aligned}
\end{equation}
The intuition behind this objective is to learn the vector field $\vz_{t}$ such that it aligns with the direction of the steepest descent (negative gradient) of the objective function. Note that the sign of the dot product matters as it determines minimizing or maximizing the property.

The proposed objective function in the supervised scenario is
\[
\gL = \gL_{r} + \gL_{\phi} + \gL_{\gP}
\]

\noindent\textbf{Unsupervised Diversity Guidance.} When no explicit potential energy function is provided to learn the flow, we need to define a potential energy function that captures the change of molecular properties. As molecular properties are determined by the structures, we devise a potential energy that maximizes the continuous structure change of the generated molecules. Inspired by~\citet{song2023latent}, we couple the traversal direction with the Jacobian of the generator to maximize the traversal variations in the molecular space. The perturbation on latent samples can be approximated by the first-order Taylor approximation:
\begin{equation}
    g(\vz_{t} + \epsilon\nabla_\vz \phi^k(t,\vz_{t}))= g(\vz_{t})+ \epsilon\frac{\partial g(\vz_{t})}{\partial \vz_{t}} \nabla_{\vz} \phi^k(t,\vz_t) + R_1(g(\vz_{t}))
    \label{eq:jvp}
\end{equation}
where $\epsilon$ denotes perturbation strength, and $R_1(\cdot)$ is the high-order terms. In the unsupervised setting, for sufficiently small $\epsilon$, if the Jacobian-vector product can cause large variations in the generated sample, the direction is likely to correspond to certain properties of molecules. We therefore introduce such a Jacobian-vector product guidance: 
\begin{equation}
    \gL_{\gJ} =-\norm*{\frac{\partial g(\vz_{t})}{\partial \vz_{t}} \nabla_\vz \phi^k(t,\vz_{t})}_2^2
\end{equation}
Compared to the supervised setting which maximizes the change of the molecular properties, it aims to find the direction that causes the maximal change of the structures. This can in turn effectively pushes the initial data distribution to the target one concentrated on the maximum property value. The Jacobian guidance will compete with the dynamical regularization (e.g. wave-like form) on the flow to yield smooth and meaningful traversal paths.

\noindent\textbf{Disentanglement Regularization.} While the above formulation can encourage smooth dynamics and meaningful output variations, the flows are likely to mine identical directions which all correspond to the maximum Jacobian change. To avoid such a trivial solution, we adopt an auxiliary classifier $l_\gamma$ following~\citet{song2023latent} to predict the flow index and use the cross-entropy loss to optimize it:
\begin{equation}
    \gL_{k}=\gL_{CE}(l_\gamma (g_\psi(\vz_{t});g_\psi(\vz_{t+1})),k)
\end{equation}
Where $\vx_{t} = g(\vz_t)$ is the generated sample from timestep $t$. We see the extra classifier guidance would encourage each flow to be independent and find distinct properties.
For each target property, we compute the Pearson correlation coefficient using a randomly generated test set.
This coefficient measures the correlation between the property and a natural sequence (from $1$ to time step $t$) along the optimization trajectory.
We then select the energy network that achieves the highest correlation score for optimizing molecules with that specific property.

The proposed objective function in the unsupervised scenario is
\[
\gL = \gL_{r} + \gL_{\phi} + \gL_{\gJ} + \gL_k
\]

\subsection{Connection with Langevin Dynamics for Global Optimization} 
In scenarios where our flow adheres to the dynamics of the Fokker-Planck equation, our approach may also be interpreted as employing a learned potential energy function to simulate Langevin Dynamics for global optimization~\cite{gardiner1985handbook}. Notably, the convergence of Langevin dynamics, particularly at low temperatures, tends to occur around the global minima of the potential energy function~\cite{chiang1987diffusion}. The continuous and discretized Langevin dynamics are as follows:
\begin{equation}
\begin{aligned}
\rd\vz_t &= -\nabla_z h_\eta(\vz_t)\rd t + \sqrt{2}\rd\mathbf{w}_t\\
\vz_t &= \vz_{t-1} - \nabla_z h_\eta(\vz_{t-1})\rd t + \sqrt{2\rd t}\mathcal{N}(0, I)
\end{aligned}
\label{eq:ld}
\end{equation}
\begin{proposition}
\label{prop:global_converge}
(Global Convergence of Langevin Dynamics, adapted from~\citet{gelfand1991recursive}). Given a Langevin dynamics in the form of 
\[\vz_t = \vz_{t-1} - a_t(\nabla_z h_\eta(\vz_{t-1}) + \vu_t) + b_t\mathbf{w}_t\]
where $\mathbf{w}_t$ is a $d$-dimensional Brownian motion, $a_t$ and $b_t$ are a set of positive numbers with $a_T, b_T \rightarrow 0$, and $\vu_t$ is a set of random variables in $\mathbb{R}^n$ denoting noisy measurements of the energy function $h_\eta(\cdot)$. Under mild assumptions, $\vz_t$ converges to the set of global minima of $h_\eta(\cdot)$ in probability.
\end{proposition}

Following \cref{prop:global_converge}, the learned latent flow can be used to search for molecules with optimal properties and it converges to the global minimizers of the learned latent potential energy function.

\section{Experiments}

\subsection{Experiment Set-up}

\paragraph{Datasets \& Molecular properties.} We extract 4,253,577 molecules from the three commonly used datasets for drug discovery including \textbf{MOSES}~\citep{Polykovskiy2018MolecularS}, \textbf{ZINC250K}~\citep{irwin2005zinc}, and \textbf{ChEMBL}~\citep{Zdrazil2023TheCD}. 
Molecules are represented by SELFIES strings~\cite{krenn2020self}. All input molecules are padded to the maximum length in the dataset before fitting into the generative model.
We consider a total of 8 molecular properties which include \textit{3 general drug-related properties} --- Quantitative Estimate of Drug-likeness~(QED), Synthesis Accessibility~(SA), and penalized Octanol-water Partition Coefficient~(plogP) and \textit{3 machine learning-based target activities} --- DRD2, JNK3 and GSK3B~\cite{huang2022artificial}
, \textit{2 simulation-based target activities} --- docking scores for two human proteins ESR1 and ACAA1.
See \cref{sec:metrics} for details.

\paragraph{Implementations.} We establish our framework by pre-training a VAE model that learns a latent space of molecules that can generate new molecules by decoding latent vectors from the latent space. We adapt the framework in \citet{eckmann2022limo} which is a basic VAE architecture with molecular SELFIES string representations and an additional MLP model as the surrogate property predictor.
See \cref{sec:exp_setup} for all implementation and hyper-parameter details.

\paragraph{Model variants.} As discussed in \cref{sec:flows}, our proposed framework is general to incorporate different dynamical priors to learn the flow. For the experiments, we consider four types of dynamics including \textit{gradient flow~(GF)}, \textit{Wave flow (Wave,~\cref{eq:wave})}, \textit{Hamilton Jacobi flow~(HJ,~\cref{eq:hj})} and \textit{Langevin Dynamics or equivalently Fokker Planck flow~(LD,~\cref{eq:ld})}.

For the specific molecular properties and evaluations, the readers are kindly referred to \Cref{sec:exp_details} for details. We also move qualitative evaluations to \Cref{sec:more_viz} due to space limit.

\subsection{Molecule Optimization}
\label{sec:mol_optim}

Molecule optimization is key in drug design and materials discovery, aiming to identify molecules with optimal properties~\cite{brown2019guacamol}.
Various machine learning methods have accelerated this process~\cite{du2022molgensurvey}.
Our discussion focuses on optimization within the latent space of generative models, primarily using gradient-based optimization as outlined in \Cref{sec:related_traversal}. 
We categorize molecule optimization into three scenarios: (1) unconstrained optimization to identify molecules with the best properties, (2) constrained optimization to find molecules with the best-expected property and similar to specific structures—a common step in the lead optimization process, and (3) multi-objective optimization to simultaneously enhance multiple properties of a molecule.

\begin{table}[htbp]
\caption{\textbf{Unconstrained plogP, QED maximization, and docking score minimization.}
(SPV denotes supervised scenarios, UNSUP denotes unsupervised scenarios).
Boldface highlights the highest-performing generation for each property within each rank.
} %
\centering
\small
\resizebox{0.9\linewidth}{!}{ %
\begin{sc}
\begin{tabular}{l|ccc|ccc|ccc|ccc}
\toprule
Method & \multicolumn{3}{c}{plogP $\uparrow$} & \multicolumn{3}{c}{QED $\uparrow$}& \multicolumn{3}{c}{ESR1 Docking  $\downarrow$} & \multicolumn{3}{c}{ACAA1 Docking $\downarrow$} \\
 & 1st & 2nd & 3rd & 1st & 2nd & 3rd & 1st & 2nd & 3rd & 1st & 2nd & 3rd \\
\midrule
Random & 3.52& 3.43& 3.37& 0.940& 0.933& 0.932 & -10.32& -10.18& -10.03& -9.86& -9.50& -9.34\\
ChemSpace & 3.74& 3.69& 3.64& 0.941& 0.936& 0.933 & \textbf{-11.66}& -10.52& -10.43& -9.81& -9.72& -9.63\\
Gradient Flow & 4.06& 3.69& 3.54& 0.944& 0.941& 0.941 & -11.00& -10.67& -10.46& -9.90& -9.64& -9.61\\
\midrule
Wave (spv)& 4.76& 3.78& 3.71& \textbf{0.947}& 0.934& 0.932 & -11.05& \textbf{-10.71}& \textbf{-10.68}& \textbf{-10.48}& \textbf{-10.04}& \textbf{-9.88}\\
 Wave (unsup)& \textbf{5.30}& \textbf{5.22}& \textbf{5.14}& 0.905& 0.902& 0.978 & -10.22& -10.06& -9.97& -9.69& -9.64& -9.57\\
HJ (spv)& 4.39& 3.70& 3.48& 0.946& 0.941& 0.940 & -10.68& -10.56& -10.52& -9.89& -9.61& -9.60\\
 HJ (unsup)& 4.26& 4.10& 4.07& 0.930& 0.928& 0.927 & -10.24& -9.96& -9.92& -9.73& -9.31& -9.24\\
LD& 4.74& 3.61& 3.55& \textbf{0.947}& \textbf{0.947}& \textbf{0.942} & -10.68& -10.29& -10.28& -10.34& -9.74& -9.64\\
\bottomrule
\end{tabular}
\end{sc}
} %
\label{tab:unconstrained_optim}
\end{table}

\paragraph{Baselines.} For molecule optimization, we follow the same experiment procedure as in \citet{eckmann2022limo}\footnote{Note that we notice there was a misalignment of normalization schemes for the plogP property in the previous literature, so we only rerun and compare with related methods that align with our normalization scheme. Details can be found in \cref{sec:metrics}.}. To ensure a fair comparison, we use the same pre-trained VAE model for all the methods. The details about the baselines are deferred to \cref{sec:baselines}. We also propose evolutionary algorithm (EA)-based latent optimization approaches in our comparison (\cref{sec:additional_baselines}).

\begin{figure}[htbp]
\begin{center}
\includegraphics[width=\linewidth]{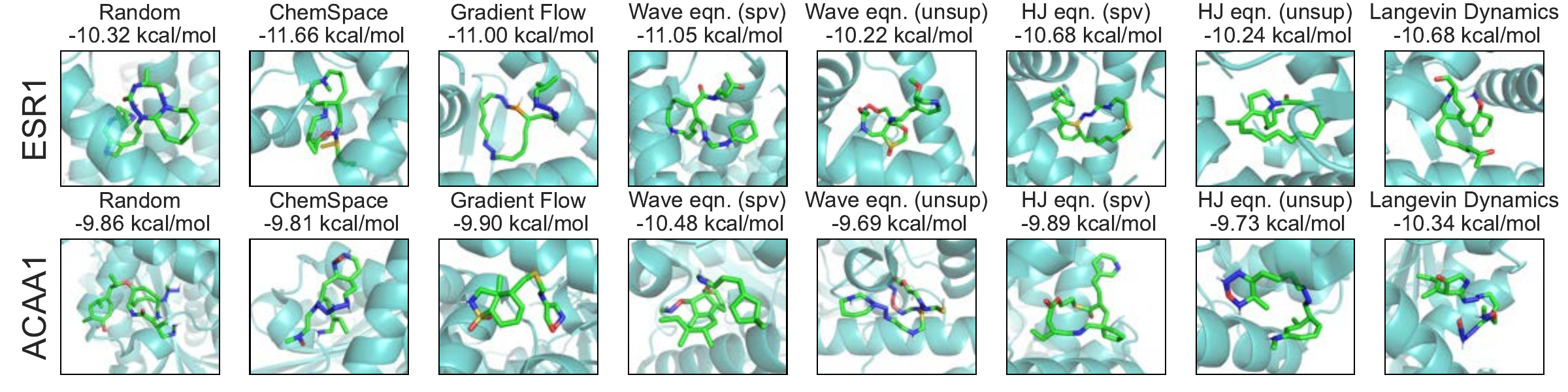}
\end{center}
\caption{
\textbf{Visualization of generated ligands docked against target ESR1 and ACAA1.}
}
\label{fig:protein_ligand}
\end{figure}

\paragraph{Unconstrained Molecule Optimization.} In this study, we randomly sample 100,000 molecules from the latent space and assess the top three scores after 10 steps of optimization for each method~(details in \Cref{tab:unconstrained_optim}). For two specific docking scores tasks, however, only 10,000 molecules are sampled due to computational resource constraints. All methods employ a step size of 0.1 to ensure a fair comparison. Our findings reveal that the efficacy of optimization methods varies with target properties, highlighting the necessity of employing a diverse set of approaches within the optimization framework, rather than depending on a single dominant method. We also visualize some generated ligands docked into protein pockets in \Cref{fig:protein_ligand}.

\begin{figure}[htbp]
\begin{center}
\includegraphics[width=\linewidth]{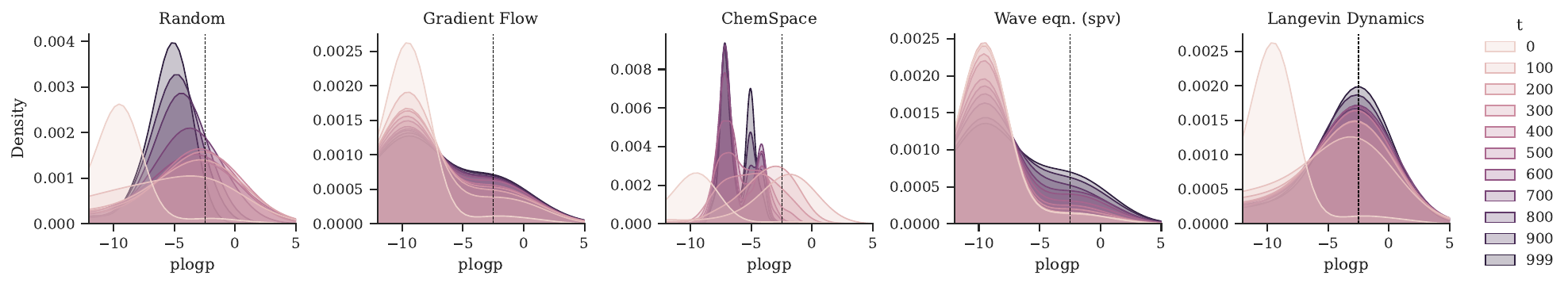}
\end{center}
\caption{\textbf{Molecular property plogP distribution shifts following the latent flow path.}
}
\label{fig:plogp_spv_kde}
\end{figure}

Furthermore, when extending the optimization procedure to 1,000 steps, illustrated in \Cref{fig:plogp_spv_kde}, Langevin dynamics significantly pushes the entire distribution to molecules with better properties, surpassing other methods in performance. Although the random direction method is effective in optimizing molecules, it does not consistently produce significant shifts in the distribution. Moreover, the outputs from ChemSpace often converge to just a few molecules, which is indicative of the challenge posed by Out-of-Distribution (OoD) generation.

\paragraph{Similarity-constrained Molecule Optimization.}
Adopting methodologies from JT-VAE~\cite{jin2018junction} and LIMO~\cite{eckmann2022limo}, we select 800 molecules with the lowest partition coefficient (plogP) scores from the ZINC250k dataset.
These molecules undergo 1,000 steps of optimization until convergence is achieved for all methods.
The optimal results for each molecule, adhering to a predefined similarity constraint ($\delta$), is reported as in \Cref{tab:optim_plogp}.
The structural similarity is measured using Tanimoto similarity between Morgan fingerprints with a radius of 2.

\begin{table}[htbp]
\caption{
\textbf{Similarity-constrained plogP maximization.}
For each method with minimum similarity constraint $\delta$, the results in reported in format mean ± standard derivation (success rate \%) of absolute improvement, where the mean and standard derivation are calculated among molecules that satisfy the similarity constraint.
} %
\centering
\resizebox{0.8\linewidth}{!}{ %
\begin{tabular}{@{}c|cccc@{}}
\toprule
Method& $\delta = 0$ & $\delta = 0.2$ & $\delta = 0.4$ & $\delta = 0.6$ \\ \midrule
Random & 11.76 ± 6.18 (99.0) & 7.64 ± 6.38 (80.0) & 5.03 ± 5.70 (52.1) & 2.37 ± 3.71 (21.1) \\
\midrule
ChemSpace & 12.13 ± 6.41 (99.8) & 9.07 ± 6.80 (90.2) & \textbf{7.52 ± 6.29 (59.4)} & \textbf{5.70 ± 5.84 (20.2)} \\
Gradient Flow & 7.88 ± 7.28 (60.4) & 7.20 ± 6.98 (56.5) & 5.45 ± 6.45 (41.9) & 3.60 ± 5.50 (18.4) \\
\midrule
Wave (spv) & 6.83 ± 7.15 (59.6) & 5.62 ± 6.42 (54.9) & 4.31 ± 5.55 (41.9) & 2.47 ± 4.21 (20.6) \\
Wave (unsup) & 19.76 ± 13.62 (99.6) & 7.47 ± 9.62 (50.2) & 2.06 ± 4.37 (27.3) & 0.77 ± 2.21 (16.8) \\
HJ (spv) & 8.58 ± 8.08 (68.0) & 6.62 ± 7.44 (60.0) & 4.27 ± 5.40 (40.6) & 2.39 ± 4.10 (18.5) \\
HJ (unsup) & \textbf{20.64 ± 12.93 (98.0)} & 8.57 ± 9.69 (50.1) & 2.12 ± 3.55 (19.5) & 0.67 ± 0.86 (8.6) \\
LD & 12.98 ± 6.23 (99.6) & \textbf{9.70 ± 6.21 (94.4)} & 6.14 ± 5.99 (70.9) & 2.94 ± 4.34 (35.4) \\
\bottomrule
\end{tabular}
} %
\label{tab:optim_plogp}
\end{table}

Without any similarity constraints, the unsupervised approaches significantly improve molecular properties with a high success rate.
However, as the similarity constraints increase, both the magnitude and success rate of unsupervised methods decrease notably.
The most considerable improvements of ChemSpace are observed when similarity constraints are set at 0.4 and 0.6.
Despite this, as shown in \Cref{fig:plogp_spv_kde} and previously discussed, the generation within ChemSpace encounters significant OoD issues after extensive iterations.
Among all techniques evaluated, Langevin dynamics stands out for its overall high improvements and high success rate. As further illustrated in \Cref{fig:optim_plogp_conv}, Langevin dynamics also demonstrates a notably faster empirical convergence rate.
We also report the performance in optimizing the QED property in \Cref{sec:more_exp}.

Surprisingly, we observe that the random direction performs well on molecule optimization tasks.
This observation motivates us to study the structure of the latent space. We show that the molecular structure distribution on the latent space follows a high-dimensional Gaussian distribution and the random direction increases the norm of the latent vectors that have strong correlations with molecular properties. We analyze this systematically in \cref{sec:latent_viz}.
It is also notable that although random directions could be effective in optimizing molecules, the distribution of the entire molecule sets being optimized does not change accordingly as shown in \cref{fig:plogp_spv_kde}.

\paragraph{Multi-objective Molecule Optimization.}
As we are learning distinct vector fields and potential energy functions for each property, they can be readily added together for multi-objective optimization~\cite{eckmann2022limo,Du2023ChemSpacEIA}.
To generate molecules that are optimized on multiple properties, we use a similar setting as similarity-constrained molecule optimization to select 800 molecules from the ZINC250k dataset with the lowest QED and aim to generate molecules with high QED as well as low SA simultaneously.
At each time step, the latent vector is optimized following the averaged direction of two corresponding flow directions.
This scheme could be seamlessly generalized to $k$-objectives optimization.
\cref{tab:multi_optimization} shows that Langevin dynamics and ChemSpace achieve the best or competitive performance at all similarity cutoff levels.

\begin{table}[htbp]
\caption{
\textbf{Similarity-constrained Multi-objective (QED-SA) maximization.}
The value of QED and SA is scaled to both have a range from 0 to 100 for an equal-weighted sum.
The method with the highest equal-weighted sum improvement of QED and SA of each structure similarity level is bolded.
} %
\centering
\small
\resizebox{0.7\linewidth}{!}{ %
\begin{tabular}{@{}c|cccc@{}}
\toprule
& $\delta = 0$ & $\delta = 0.2$ & $\delta = 0.4$ & $\delta = 0.6$ \\ \midrule
& \multicolumn{4}{c}{\textbf{QED}} \\ \midrule
Random & 45.5 ± 13.3 (99.5) & 20.7 ± 14.4 (81.8) & 12.8 ± 10.7 (57.0) & 8.0 ± 8.3 (29.5) \\
ChemSpace & 47.0 ± 12.9 (99.6) & 25.9 ± 16.7 (88.2) & \textbf{15.5 ± 13.2 (63.4)} & \textbf{9.7 ± 10.1 (31.6)} \\
Gradient Flow & 31.9 ± 17.9 (89.9) & 23.0 ± 16.4 (80.0) & 14.4 ± 12.6 (59.4) & 9.4 ± 9.1 (32.2) \\
Wave (SPV) & 14.6 ± 14.4 (26.8) & 12.3 ± 11.7 (24.2) & 9.8 ± 10.0 (19.8) & 6.8 ± 6.5 (12.4) \\
Wave (UNSUP) & 39.0 ± 18.5 (96.1) & 18.3 ± 14.0 (69.4) & 10.0 ± 9.6 (43.2) & 6.4 ± 7.0 (24.9) \\
HJ (SPV) & 45.2 ± 13.7 (98.9) & 22.7 ± 15.6 (84.9) & 13.9 ± 12.0 (57.5) & 9.3 ± 9.7 (30.2) \\
HJ (UNSUP) & 40.6 ± 19.4 (96.8) & 15.5 ± 12.8 (71.4) & 9.2 ± 9.4 (46.2) & 6.4 ± 7.3 (26.9) \\
LD & \textbf{47.0 ± 13.1 (99.6)} & \textbf{27.6 ± 16.3 (92.1)} & 15.4 ± 12.5 (71.4) & 9.6 ± 9.4 (40.1) \\ \midrule
& \multicolumn{4}{c}{\textbf{SA}} \\ \midrule
Random & 8.34 ± 8.06 (37.2) & 6.70 ± 7.11 (27.0) & 4.80 ± 5.73 (19.1) & 2.61 ± 3.21 (10.9) \\
ChemSpace & 7.92 ± 7.71 (42.8) & 6.63 ± 6.71 (36.1) & \textbf{4.75 ± 5.45 (25.2)} & \textbf{2.89 ± 3.22 (15.4)} \\
Gradient Flow & 9.56 ± 8.49 (44.0) & 7.19 ± 6.66 (35.8) & 5.27 ± 5.30 (26.6) & 3.04 ± 3.62 (17.1) \\
Wave (SPV) & 6.37 ± 6.30 (12.9) & 5.77 ± 5.81 (12.4) & 4.54 ± 4.51 (10.5) & 3.44 ± 3.59 (7.1) \\
Wave (UNSUP) & 15.21 ± 10.17 (89.2) & 7.69 ± 6.51 (70.8) & 3.92 ± 3.86 (45.9) & 2.22 ± 1.97 (25.9) \\
HJ (SPV) & 8.93 ± 8.39 (52.1) & 6.69 ± 6.61 (38.8) & 5.21 ± 5.72 (29.2) & 3.23 ± 3.40 (18.6) \\
HJ (UNSUP) & 16.03 ± 10.31 (91.0) & 7.35 ± 6.34 (68.5) & 4.22 ± 4.09 (46.8) & 2.73 ± 2.85 (27.8) \\
LD & \textbf{11.51 ± 10.44 (69.2)} & \textbf{7.51 ± 7.41 (45.4)} & 4.50 ± 4.95 (33.6) & 2.75 ± 3.09 (20.1) \\
\bottomrule
\end{tabular}
} %
\label{tab:multi_optimization}
\end{table}

\subsection{Molecule Manipulation}

Molecule manipulation is a relatively new task proposed in \citet{Du2023ChemSpacEIA} to study the performance of latent traversal methods. Specifically, the main idea of molecule manipulation is to find smooth local changes of molecular structures that simultaneously improve molecular properties which is essential to help chemists systematically understand the chemical space.

\begin{table}[htbp]
\caption{
\textbf{Success Rate of traversing latent molecule space to manipulate over a variety of molecular properties.}
Numbers reported are strict success rate/relaxed success rate in \%. (SPV denotes supervised scenarios, UNSUP denotes unsupervised scenarios). The ranking is the average between the ranking of average strict success rate and ranking of the average relaxed success rate. 
}
\centering
\small
\resizebox{.95\linewidth}{!}{ %
\begin{sc}
\begin{tabular}{c|c|c|cccccc}
\toprule
 & Ranking & Average & plogP ($\uparrow$) & QED ($\uparrow$) & SA ($\downarrow$) & DRD2 ($\uparrow$) & JNK3 ($\uparrow$) & GSK3B ($\uparrow$)  \\
\midrule
Random-1D & 9 & 1.42 / 6.85 &6.00 / 31.60&0.00 / 0.00&0.00 / 0.00&0.00 / 0.00&2.50 / 9.50&0.00 / 0.00 \\
Random & 6 & 0.57 / 42.3 &0.00 / 32.60&0.10 / 3.20&0.40 / 8.60&0.50 / 87.10&1.50 / \textbf{81.40}&0.90 / 40.90 \\ \midrule
ChemSpace & 3 & 6.17 / 22.83 &5.20 / 25.00&6.00 / \textbf{18.10}&\textbf{6.80} / \textbf{26.50}&3.20 / 18.00&8.60 / 25.80&\textbf{7.20} / 23.60 \\ \midrule
Wave (unsup) & 3 & 1.18 / \textbf{45.28} &0.60 / \textbf{40.30}&0.60 / 6.20&1.90 / 16.50&0.40 / 86.40&1.80 / 78.20&1.80 / \textbf{44.10} \\
Wave (spv) & 7 & 1.85 / 8.08 &0.00 / 0.20&3.40 / 12.10&4.00 / 18.60&3.50 / 17.10&0.00 / 0.20&0.20 / 0.30 \\
HJ (unsup) & 3 &  2.3 / 25.28 &3.00 / 15.60&0.70 / 3.20&1.70 / 13.00&0.20 / \textbf{87.20}&4.80 / 18.70&3.40 / 14.00 \\
HJ (spv) & 7 &1.97 / 7.4 &3.00 / 13.20&3.00 / 7.20&3.70 / 15.00&1.90 / 8.50&0.20 / 0.50&0.00 / 0.00 \\
GF (spv) & 1 & \textbf{7.62} / 28.78 &\textbf{6.90} / 28.30&\textbf{6.60} / 16.70&6.30 / 25.10&7.10 / 36.10&\textbf{11.70} / 35.50&7.10 / 31.00 \\
LD (spv) & 2 & 6.23 / 26.68 &5.90 / 26.00&6.20 / 15.50&5.20 / 22.90&\textbf{6.00} / 33.30&8.40 / 33.80&5.70 / 28.60 \\
\bottomrule
\end{tabular}
\end{sc}
} %
\label{tab:traversal}
\end{table}

\textbf{Supervised Molecule Manipulation}
\cref{tab:traversal} shows the success rate results of manipulating 1,000 randomly sampled molecules to optimize each desired property. Following \citet{Du2023ChemSpacEIA}, we traverse the latent space for 10 steps in the traversal direction of each method and report the strict and relaxed success rate.
Details of the definition of these metrics can be found in \cref{sec:eval_metrics}.
Among all approaches, the gradient flow achieves the highest success rates in multiple properties such that it takes the steepest descent of the surrogate model. When the step size is small enough, it is reasonable to learn a smooth path. However, the results still vary across properties.

\textbf{Unsupervised Molecule Manipulation}
As the correspondence between specific molecular properties and learned latent flows is not explicitly given in the unsupervised scenario, we use an artificial process to mimic the use case in reality.
Specifically, we learn 10 different potential energy functions representing 10 disentangled flows following \cref{alg:train} using Wave equation and Hamilton-Jacobi equation and validated them on 1,000 unseen molecules.
For each flow, we evaluate the properties of molecules generated along the 10-step manipulation trajectory.
The learned potential energy function with the highest correlation score is selected for each property representing the learned Jacobian structural change that would most effectively optimize the corresponding property.

In \cref{tab:traversal}, we can observe that even though it is without supervised training of traversal directions, the flow still learns meaningful directions from molecular structure to property changes.
Surprisingly, the relaxed success rate of manipulating molecules for JNK3, and GSK3B in unsupervised settings are better than in supervised settings.
We hypothesize that this is partially because of the training and generalization errors of the surrogate model. On the contrary, the structure change measurement does not provide supervision but is correlated with key molecular properties. We would like to point out that this is an open question in chemistry, often referred as to the structure-activity relationship~\cite{dudek2006computational}, such that it is important to know the correspondence between structure and activity. We believe this is a promising result to demonstrate that generative models ``realize" molecular property by learning from structures. 

Among the quantitative results, it is interesting that the random direction achieves a good relaxed success rate for some properties, we argue this is because of the specific property of the learned latent space.
The latent space learned by generative models tend to be smooth such that similar molecular structures are often mapped to close areas in the latent space.
In \cref{sec:latent_viz}, we find that some molecular properties are highly correlated with their latent vector norms, in which a random direction always increases the norm and thus successfully manipulates a portion of molecules by chance.

\section{Conclusion, Limitation and Future Work}
\label{sec:conclusion}

In this paper, we propose a unified framework for navigating chemical space through the learned latent space of molecule generative models. Specifically, we formulate the traversal process as a flow that defines a vector to transport the mass of molecular distribution through time to desired concentrations (e.g. high properties). Two forces (supervised potential guidance and unsupervised structure diversity guidance) are derived to drive the dynamics. We also propose a variety of new physical PDEs on the dynamics which exhibit different properties. We hope this general framework can open up a new research avenue to study the structure and dynamics of the latent space of molecule generative models.

\textbf{Limitation and future work.} This work is a preliminary study on small molecules and it may be interesting to see it transfer to larger molecular systems or more specialized systems and properties. Beyond molecules, this approach has the potential to be extended to languages and other data modalities.

\section{Acknowledgement}
Y.D. would like to thank Ziming Liu and Kirill Neklyudov for helpful discussions.

{
    \small

    \renewcommand{\bibname}{References}
    \bibliographystyle{abbrvnat}
    \bibliography{main}
}

\clearpage
\appendix

\begin{center}
	{\Large \textbf{Appendix for ChemFlow}}
\end{center}

\startcontents[sections]
\printcontents[sections]{l}{1}{\setcounter{tocdepth}{2}}
\newpage

\section{Wasserstein Gradient Flow}
\label{sec:app_wasserstein}
As shown in the main paper, based on the dynamic formulation of optimal transport~\citep{benamou2000computational}, the $L_2$ Wasserstein distance can be re-written as:
\begin{equation}
    W_{2}(\mu_{0},\mu_{1})=\min_{\rho,v}\sqrt{\int\int \rho_{t}(\vz)|v_{t}(\vz)|^2 \mathop{d\vz dt}}
    \label{eq_w2}
\end{equation}
where $v_{t}(\vz)$ is the velocity of the particle at position $\vz$ and time $t$, and $\rho_{t}(\vz)$ is the density $\mathop{d\mu(\vz)}=\mathop{\rho_{t}(\vz)d\vz}$. The distance can be optimized by the gradient flow of a certain function on space and time. Consider the functional $\gF: \mathbb{R}^{n}\rightarrow\mathbb{R}$ that takes the following form:
\begin{equation}
    \gF(\mu) = \int U(\rho_{t}(\vz))\mathop{d\vz}
\end{equation}
The curve is considered as a gradient flow if it satisfies $\nabla \gF =  - \frac{\mathop{d}}{\mathop{dt}} \rho_{t}(\vz)$~\citep{ambrosio2005gradient}. Moving the particles leads to:
\begin{equation}
    \frac{\mathop{d}}{\mathop{dt}} \gF(\mu) = \int U'(\vz)\frac{\mathop{d}\rho_{t}(\vz)}{\mathop{dt}}\mathop{d\vz}
    \label{eq:F_d}
\end{equation}
The velocity vector satisfies the continuity equation:
\begin{equation}
   \frac{\mathop{d}\rho_{t}(\vz)}{\mathop{dt}} = -\nabla\cdot\Big(v_{t}(\vz)\rho_{t}(\vz)\Big)
\end{equation}
where $-\nabla\cdot\Big(v_{t}(\vz)\rho_{t}(\vz)\Big)$ is the tangent vector at point $\rho_{t}(\vz)$. \Cref{eq:F_d} can be simplified to:
\begin{equation}
\begin{aligned}
    \frac{\mathop{d}}{\mathop{dt}} \gF(\mu) &= \int - U'(\rho_{t}(\vz))\nabla\cdot\Big(v_{t}(\vz)\rho_{t}(\vz)\Big)\mathop{d\vz} \\
    &= \int\nabla\Big(U'(\rho_{t}(\vz))\Big)v_{t}(\vz)\rho_{t}(\vz)\mathop{d\vz}\\
    \label{eq:tang_vec}
\end{aligned}
\end{equation}
On the other hand, the calculus of differential geometry gives
\begin{equation}
    \frac{\mathop{d}}{\mathop{dt}} \gF(\mu) = {\rm Diff}\gF|_{\rho_{t}}(-\nabla\cdot\Big(v_{t}(\vz)\rho_{t}(\vz)\Big)) =\langle \nabla\gF, -\nabla\cdot\Big(v_{t}(\vz)\rho_{t}(\vz)\Big)\rangle_{f} 
\end{equation}
where $\langle,\rangle_{f}$ is a Riemannian distance function which is defined as:
\begin{equation}
    \langle -\nabla\cdot\Big(w_{1}(\vz)\rho_{t}(\vz)\Big),-\nabla\cdot\Big(w_{2}(\vz)\rho_{t}(\vz)\Big)\rangle_{f} = \int w_{1}(\vz)w_{2}(\vz) f(\vz)\mathop{d\vz}
\end{equation}
This scalar product coincides with the $W_{2}$ distance according to~\citet{benamou2000computational}. Then \cref{eq:tang_vec} can be similarly re-written as:
\begin{equation}
    \frac{\mathop{d}}{\mathop{dt}} \gF(\mu) = \langle  - \nabla\cdot\Big( \nabla U'(\rho_{t}(\vz))\rho_{t}(\vz) \Big) , -\nabla\cdot\Big(v_{t}(\vz)\rho_{t}(\vz)\Big)\rangle_{f}
\end{equation}
So the relation arises as:
\begin{equation}
\begin{aligned}
    \nabla \gF &=  - \nabla\cdot\Big( \nabla U'(\rho_{t}(\vz))\rho_{t}(\vz) \Big) 
\end{aligned}
\end{equation}

Since we have $\nabla \gF =  - \frac{\mathop{d}}{\mathop{dt}} \rho_{t}(\vz)$, the above equation can be re-written as 
\begin{equation}
\frac{\mathop{d}}{\mathop{dt}} \rho_{t}(\vz) =   \nabla\cdot \Big(\nabla U'(\rho_{t}(\vz))\rho_{t}(\vz)\Big)
\end{equation}

The above derivations can be alternatively made by the poineering JKO scheme~\citep{Jordan1996THEVF}. This explicitly defines the relation between evolution PDEs of $\rho_{t}(\vz)$ and the internal energy $U$. For our method, we use the gradient of our scalar energy field $\nabla u(\vz,t)$ to learn the velocity field which is given by $U'(\rho_{t}(\vz))$. Interestingly, driven by certain specific velocity fields $\nabla u(\vz,t)$, the evolution of $\rho(\vz,t)$ would become some special PDEs. Here we discuss some possibilities:

\noindent\textbf{Heat Equations.} If we consider the energy function $U$ as the weighted entropy:
\begin{equation}
    U(\rho_{t}(\vz)) = \rho_{t}(\vz)\log(\rho_{t}(\vz)) 
\end{equation}
We would have exactly the heat equation:
\begin{equation}
\frac{\mathop{d}}{\mathop{dt}} \rho_{t}(\vz) - \frac{\mathop{d}}{\mathop{d\vz^2}} \rho_{t}(\vz) = 0
\end{equation}
Injecting the above equation back into the continuity equation leads to the velocity field $v_{t}(\vz)$ as
\begin{equation}
\begin{aligned}
    \frac{\mathop{d}\rho_{t}(\vz)}{\mathop{dt}} &= -\nabla\cdot\Big(v_{t}(\vz)\rho_{t}(\vz)\Big) =  \frac{\mathop{d}}{\mathop{d\vz^2}} \rho_{t}(\vz)\\
    v_{t}(\vz) &= - \frac{\nabla \rho_{t}(\vz)}{\rho_{t}(\vz)} = - \nabla \log(\rho_{t}(\vz))
\end{aligned}
\end{equation}
When our $\nabla u(\vz,t)$ learns the velocity field $- \nabla \log(\rho_{t}(\vz))$, the evolution of $\rho(\vz,t)$ would become heat equations.

\noindent\textbf{Fokker Planck Equations.} For the energy function defined as:
\begin{equation}
    U(\rho_{t}(\vz)) = - A \cdot \rho_{t}(\vz) + \rho_{t}(\vz)\log(\rho_{t}(\vz))
\end{equation}
we would have the Fokker-Planck equation as
\begin{equation}
    \frac{\mathop{d}}{\mathop{dt}} \rho_{t}(\vz) + \frac{\mathop{d}}{\mathop{d\vz}} [\nabla A \rho_{t}(\vz)] - \frac{\mathop{d}}{\mathop{d\vz^2}} [\rho_{t}(\vz)]= 0, 
\end{equation}
The velocity field can be similarly derived as
\begin{equation}
    \begin{gathered}
    v_{t}(\vz) = \nabla A - \nabla \log(\rho_{t}(\vz))
    \end{gathered}
\end{equation}
For the velocity field $\nabla A - \nabla \log(\rho_{t}(\vz))$, the movement of $\rho(\vz,t)$ is the Fokker Planck equation.

\noindent\textbf{Porous Medium Equations.} If we define the energy function as
\begin{equation}
    U(\rho_{t}(\vz)) = \frac{1}{m-1} \rho_{t}^{m}(\vz)
\end{equation}
Then we would have the porous medium equation where $m>1$ and the velocity field:
\begin{equation}
    \frac{\mathop{d}}{\mathop{dt}} \rho_{t}(\vz) - \frac{\mathop{d}}{\mathop{d\vz^2}} \rho_{t}^{m}(\vz) = 0,\ v_{t}(\vz) = -m \rho^{m-2} \nabla\rho
\end{equation}
When the $\nabla u(\vz,t)$ learns the velocity $-m \rho^{m-2} \nabla\rho$, the trajectory of $\rho(\vz,t)$ becomes the porous medium equations.

\section{PDE-regularized Latent Space Learning}
Our framework can be extended to incorporate the PDE dynamics as part of the training procedure to encourage a more structured representation. To validate the effectiveness, we incorporate a PDE loss such that we expect any path in the latent space to follow specific dynamics (wave equation in our experiment).
In addition to the initial setup outlined in \Cref{sec:exp_setup}, we further fine-tune the VAE model by applying a PDE regularization loss term, defined as
\[
\gL = \gL_{VAE} + \gL_{r} + \gL_{\phi}
\]

that includes the velocity-matching objective 
$\gL_{r}$ and boundary condition $\gL_{\phi}$.
This PDE-regularized latent space learning can also be adapted to other generative models by replacing $\gL_{VAE}$.
We further optimize the model and energy network for 10 epochs across the full training dataset using an AdamW optimizer with a 1e-4 learning rate and a cosine learning rate scheduler, without a warm-up period.
All other training parameters remain consistent with those initially described in \Cref{sec:exp_setup}.

\begin{table}[htbp]
\caption{\textbf{Single-objective Maximization with PDE-regularized Latent Space Learning}
The results are the same as of \Cref{tab:unconstrained_optim}, only unsupervised results are presented for fair comparison.
(UNSUP denotes unsupervised scenarios).
$K$ represents the number of energy networks trained for unsupervised training with disentanglement regularization.
} %
\centering
\small
\resizebox{0.62\linewidth}{!}{ %
\begin{sc}
\begin{tabular}{l|c|ccc|ccc}
\toprule
Method  & $K$ & \multicolumn{3}{c}{plogP $\uparrow$} & \multicolumn{3}{c}{QED $\uparrow$} \\
 && 1st & 2nd & 3rd & 1st & 2nd & 3rd \\
\midrule
Random  &N$/$A& 3.52& 3.43& 3.37& \textbf{0.940}& \textbf{0.933}& 0.932 \\
 Wave (unsup) &10& \textbf{5.30}& \textbf{5.22}& \textbf{5.14}& 0.905& 0.902& 0.978 \\
 HJ (unsup) &10& 4.26& 4.10& 4.07& 0.930& 0.928& 0.927 \\
\midrule
Wave (unsup FT) &1& 3.71& 3.58& 3.46& 0.936& \textbf{0.933}& \textbf{0.933}\\
\bottomrule
\end{tabular}
\end{sc}
} %
\label{tab:uc_finetune}
\end{table}

During fine-tuning, the VAE loss drops from 0.2187 to 0.06288.
The results of QED optimization surpass those of two other unsupervised methods, indicating potential areas for future research and refinement.

\section{Extended Related Work}
\label{sec:related_work}

\subsection{Machine Learning for Molecule Generation}
Molecules are highly discrete objects and two branches of methods are thus developed to design or search new molecules~\cite{du2022molgensurvey}. One idea is to leverage the advancement of deep generative models which approximate the data distribution from a provided dataset of molecules and then sample new molecules from the learned density. This idea inspires a line of work developing deep generative models from variational auto-encoders~(VAE)~\cite{gomez2018automatic,jin2018junction}, generative adversarial networks~(GAN)~\cite{guimaraes2017objective,cao2018molgan}, normalizing flows~(NF)~\cite{madhawa2019graphnvp,zang2020moflow} and more recently diffusion models~\cite{hoogeboom2022equivariant,vignac2023digress,jo2022score}. However, to respect the combinatorial nature of molecules, another line of work leverage combinatorial optimization to search new molecules including genetic algorithm~(GA)~\cite{jensen2019graph}, Monte Carlo tree search~(MCTS)~\cite{yang2017chemts}, reinforcement learning~(RL)~\cite{You2018-xh}, but often with sophisticated optimization objectives beyond simple valid molecules.

\subsection{Goal-oriented Molecule Generation}
In addition to simply generating valid molecules, a more realistic application is to generate molecules with desired properties~\cite{du2022molgensurvey}. For deep generative model-based methods, it is naturally combined with on-the-fly optimization methods such as gradient-based or Bayesian optimization (in low data regime) as it often maps data to a low-dimensional and smooth latent space thus more friendly for these optimization methods~\cite{griffiths2020constrained}. For methods that do not explicitly reduce the dimensionality of data such as diffusion models, \citet{diffsbdd} propose an evolutionary process to iteratively optimize the generated molecules. As it is observed that the learned latent space exhibits explicit structure~\cite{gomez2018automatic}, \citet{Du2023ChemSpacEIA} leverage such property to learn a linear classifier to find the latent direction to optimize the property of given molecules. In opposition to deep generative models, combinatorial optimization methods are often inherently associated with optimization, e.g. reward function in RL, selection criteria in GA, etc~\cite{fu2022reinforced,loeffler2023reinvent4}.

\subsection{Image Editing in the Latent Space}
\label{sec:traversal_img_edit}
Beyond molecule generation, there is a vast literature on the study of the latent space of generative models on images for image editing and manipulation~\cite{goetschalckx2019ganalyze,jahanian2020steerability,voynov2020unsupervised,harkonen2020ganspace,zhu2020learning,peebles2020hessian,shen2021closed,song2022orthogonal,song2023latent,song2023flow,song2023householder,song2024unsupervised}. Here we highlight some representative supervised and unsupervised approaches. Supervised methods usually require pixel-wise annotations. InterfaceGAN~\citep{shen2020interfacegan} leverages face image pairs of different attributes to interpret disentangled latent representations of GANs. \citet{jahanian2020steerability} explores linear and non-linear walks in the latent space under the guidance of user-specified transformation. Compared to supervised methods, unsupervised ones mainly focus on discovering meaningful interpretable directions in the latent space through extra regularization. \citet{voynov2020unsupervised} proposes to jointly learn a set of orthogonal directions and a classifier to learn the distinct interpretable directions. SeFa~\cite{shen2021closed} and HouseholderGAN~\cite{song2023householder} propose to use the eigenvectors of the (orthogonal) projection matrices as interpretable directions to traverse the latent space. More relevantly, \citet{song2023latent} proposes to use wave-like potential flows to model the spatiotemporal dynamics in the latent spaces of different generative models.

\section{Experiments Details}
\label{sec:exp_details}

\subsection{Baselines}
\label{sec:baselines}

We compare with the following baselines:
\begin{itemize}
    \item \textbf{Random}: we take a linear direction that is sampled from Multi-variant Gaussian distribution in the high dimensional latent space and normalized to unit length for all molecules across all time steps.
    \item \textbf{Random 1D}: we take a unit vector where only 1 randomly selected dimension is either 1 or -1 as the linear direction.
    \item \textbf{ChemSpace}~\cite{Du2023ChemSpacEIA}: a separation boundary of the training dataset in latent space w.r.t. the desired property is classified by an Support vector machine~(SVM).
    Then we take the normal vector corresponding to the positive separation as the manipulation direction of control.
    \item \textbf{Gradient Flow} \textbf{(LIMO)}~\cite{eckmann2022limo}: a VAE-based generative model that encodes the input molecules into SELFIES~\cite{Krenn2019SelfreferencingES} and auto-regressive on the tokenized molecule.
    LIMO uses Adam optimizer to reverse optimize on the input latent vector $z$ whereas Gradient Flow is equivalent to using an SGD optimizer for the same purpose.
    \item \textbf{Evolutionary Algorithm (EA)}: EA is a general framework that leverages random mutation and crossover to select better samples iteratively to search good solutions. In our scenario, we realize an EA-based baseline by perturbing the vector by the directions given by random, ChemSpace and gradient. The pseudocode can be found in \Cref{alg:ea}. The results are presented at \Cref{sec:additional_baselines}.
\end{itemize}

\subsection{Training Dataset}

\begin{itemize}[nosep]
    \item \textbf{ChEMBL}~\citep{Zdrazil2023TheCD} is a database of 2.4M bioactive molecules with drug-like properties, including features like a Natural Product likeness score and annotations for chemical probes and bioactivity measurements.
    \item \textbf{MOSES}~\citep{Polykovskiy2018MolecularS} is a benchmarking dataset derived from the ZINC~\citep{irwin2005zinc} Clean Leads collection, containing 2M filtered molecules with specific physicochemical properties, organized into training, test, and unique scaffold test sets to facilitate the evaluation of model performance on novel molecular scaffolds.
    \item \textbf{ZINC250k} is a subset of the ZINC database containing $\sim$250,000 commercially available compounds for virtual screening.
\end{itemize}

\subsection{Molecule Properties}
\label{sec:metrics}
We report the following metrics for our experiments:
\begin{itemize}
    \item \textbf{Penalized logP/plogP}: Estimated octanol-water partition coefficient penalized by synthetic accessibility~(SA) score and the number of atoms in the longest ring.
    \item \textbf{QED}: Quantitative Estimate of Drug-likeness, a metric that evaluates the likelihood of a molecule being a successful drug based on its pharmacophores and physicochemical properties.
    \item \textbf{SA}: Synthetic Accessibility, a score that predicts the ease of synthesis of a molecule, with lower values indicating easier synthesis.
    \item \textbf{DRD2 activity}: Predicted activity against the D2 dopamine receptor, using machine learning models trained on known bioactivity data.
    \item \textbf{JNK3 activity}: Predicted activity against the c-Jun N-terminal kinase 3, important for developing treatments for neurodegenerative diseases.
    \item \textbf{GSK3B activity}: Predicted activity against Glycogen Synthase Kinase 3 beta, which plays a crucial role in various cellular processes including metabolism and neuronal cell development.
    \item \textbf{ESR1 docking score}: Simulation-based score representing the binding affinity of a molecule to Estrogen Receptor 1, relevant in the context of breast cancer therapies.
    \item \textbf{ACAA1 docking score}: Simulation-based score representing the binding affinity of a molecule to Acetyl-CoA Acyltransferase 1, important for metabolic processes in cells.
\end{itemize}

\subsubsection{Misalignment of normalization schemes for penalized logP}
\label{sec:norm_plogp}
We notice that plogP is a commonly reported metric in recent molecule discovery literature but does not share the same normalization scheme.
Following \citet[Eq. 1]{gomez2018automatic}, the SA scores and a ring penalty term were introduced into the calculation of penalized logP as the following
\[
J^{\logp}(m) = \logp(m)-\sa(m)-\text{ring-penalty}(m)
\]
Each term of $\logp(m)$, $\sa(m)$, and $\text{ring-penalty}(m)$ are normalized to have zero mean and unit standard derivation across the training data.
However, no sufficient details were included in their paper or their released source code on how the $\text{ring-penalty}(m)$ is computed.
Specifically, 3 implementations are widely used in various works.

\paragraph{Penalized by the length of the maximum cycle without normalization}
$\text{ring-penalty}(m)$ is computed as the number of atoms on the longest ring minus 6 in their implementation.
Neither $\logp(m)$, $\sa(m)$, or $\text{ring-penalty}(m)$ is normalized.
MolDQN~\citep{Zhou2018OptimizationOM} reported their results in this scheme.

\paragraph{Penalized by the length of the maximum cycle with normalization}
$\text{ring-penalty}(m)$ is computed same as without normalization.
MARS~\citep{Xie2021MARSMM}, HierVAE~\citep{Jin2020HierarchicalGO}, GCPN~\citep{You2018GraphCP}, and ours report plogP using this scheme.

\paragraph{Penalized by number of cycles} As described by \citet[page 7 footnote 3]{jin2018junction}, $\text{ring-penalty}(m)$ is computed as the number of rings in the molecule that has more than 6 atoms.
LIMO reports plogP using this metric.

\subsection{Training and inference}

We detail our training and inference workflows in Alg.~\ref{alg:train} and Alg.~\ref{alg:inference}, respectively.

\label{sec:train_inference_algo}
\begin{algorithm}
\caption{ChemFlow Training}\label{alg:train}
\begin{algorithmic}[1]
\REQUIRE Pre-trained encoder $f_\theta$, decoder $g_\psi$, (optional) classifier $l_\gamma$, timestamps $T$, \# of potential functions $K$
\STATE Initialize $\phi^j(\cdot)\gets$ MLP for $j=1,\dots,K$
\REPEAT
\STATE Sampling: $\vz_0 = f_\theta(x_0)$, $t\sim \texttt{Categorical}(T)$, $k\sim \texttt{Categorical}(K)$\\
\FOR{$i = 1,\dots,t$}
\STATE$\vz_{i+1} = \vz_{i} + \nabla_\vz \phi^k(i,\vz_{i})$
\ENDFOR
\STATE Decode: $\vx_t = g_\psi(\vz_t)$, $\vx_{t+1} = g_\psi(\vz_{t+1})$
\IF{unsupervised}
\STATE Classification: $ \hat{k}= l_\gamma (\vx_{t};\vx_{t+1})$
\STATE Loss: $\gL = \gL_{r} + \gL_{\phi} + \gL_{\gJ} + \gL_k$ 
\ELSE
\STATE Loss: $\gL = \gL_{r} + \gL_{\phi} + \gL_{\gP}$
\ENDIF
\STATE Back-propagation through the Loss $\gL$
\UNTIL Convergence
\end{algorithmic}
\end{algorithm}

\begin{algorithm}
\label{alg:inference}
\caption{ChemFlow Inference / Traversal}
\begin{algorithmic}[1]
\REQUIRE Pre-trained encoder $f_\theta$, pre-trained potential function $\phi$, (optional) pre-trained proxy function $h$, timestamps $T$, step size $\alpha$, LD strenth $\beta$
\STATE Sampling: $\vz_0 = f_\theta(x_0)$
\FOR{$t=1,\dots,T$}
\IF{Langevin Dynamics}
\STATE $\vz_t=\vz_{t-1}-\alpha\nabla_\vz h_\eta(\vz_{t-1})+\beta\sqrt{2\alpha}\gN(0,I)$
\ELSIF{Gradient Flow}
\STATE $\vz_t=\vz_{t-1}-\alpha\nabla_\vz h_\eta(\vz_{t-1})$
\ELSE
\STATE $\vz_t=\vz_{t-1}+\alpha\nabla_\vz \phi(t-1,\vz_{t-1})$
\ENDIF
\ENDFOR
\end{algorithmic}
\end{algorithm}

\subsection{Experiments Setup}
\label{sec:exp_setup}

\paragraph{Pre-trained VAE} We follow the VAE architecture from LIMO consisting of a 128 dimension embedding layer, 1024 latent space size, 3-hidden-layer encoder, and 3-hidden-layer decoder both with 1D batch normalization and non-linear activation functions.
The hidden layer sizes are $\{4096,2048,1024\}$ for the encoder and reversely for the decoder.
We empirically find that replacing the ReLU activation function with its newer variant Mish activation function~\citep{Misra2020MishAS} results in faster convergence and better validation loss.
All the experiments reported in this paper use this Mish-activated variant of VAE.

The VAE is trained using an AdamW~\citep{Loshchilov2017DecoupledWD} optimizer, 0.001 initial learning rate, and 1,024 training batch size.
To better prevent the model from being stacked at a sub-optimal local minimum, a cosine learning rate scheduler with a 1e-6 minimum learning rate with periodic restart is applied.
The VAE is trained for 150 epochs with 4 restarts on 90\% of the training data and validated with the rest 10\% data.
The checkpoint corresponding to the epoch with the lowest validation loss is selected.
Training 150 epochs takes $\sim$8 hours on a single RTX 3090 desktop.

\paragraph{Surrogate Predictor}
The performance of the surrogate predictor is crucial to the proposed latent traversal framework.
To handle chemical properties of different magnitudes, we normalize all chemical properties in the training data to have zero mean and unit variance.
Then we use a pre-activation-norm MLP with residual connections as the surrogate predictor.
The predictor contains 3 residual blocks of size 1024 and the output dimension is 1.
Similar to the LIMO setups, we find that the choice of optimizer and training hyperparameters like learning rate or learning rate scheduler is crucial for successful training.
The predictor is trained for 20 epochs on 100,000 randomly generated samples and validated with 10,000 unseen data with SGD optimizer, 0.001 learning rate, and batch size 1000.
The epoch with the best validation loss is selected.
Training each predictor takes less than a minute.

\paragraph{Energy Network} We use an MLP structure to parameterize the energy function (the spatial derivative gives the velocity).
The time input $t$ is embedded with a sinusoidal positional embedding followed by a linear layer.
The special input $x$ is encoded with a linear layer and ReLU activation function.
The training of the network uses 9,000 random data and 1,000 unseen data for validation.
For unsupervised settings, 10 disentangled potential energy functions are trained for 310 epochs with a batch size of 100.
The epoch with the best validation loss is selected.
Training an energy network with 10 disentangled potential energy functions takes $\sim$40 minutes.

\paragraph{Reproducibility}
All the experiments including baselines are conducted on one RTX 3090 GPU and one Nvidia A100 GPU.
All docking scores are computed on one RTX 2080Ti GPU.
The code implementation is available at 
\href{https://github.com/garywei944/ChemFlow}{https://github.com/garywei944/ChemFlow}.

\subsection{Evaluation Metrics}
\label{sec:eval_metrics}

\paragraph{Success Rate} The success rate is used as the evaluation metric for the molecule manipulation task. It first randomly generates $n$ molecules and traverses each of them in the latent space for $k$ steps. The success rate is calculated as the percentage of $k$-step trajectories that are successful. In our case, we generate 1000 molecules and traverse for 10 steps. The manipulation is successful if the local change in molecular structure is smooth and molecular property is increased. Specifically, we showed two success rates: the strict success rate and the relaxed success rate. 

For the strict success rate, manipulation is a success if the molecular property is monotonically increasing, molecular similarity with respect to the previous step is monotonically decreasing, and molecules are diversity on the manipulation trajectory. These constraints are formulated as follows: 

\begin{equation}
\begin{aligned}
    C_{SP}(x,k,P) &= \mathbf{1}[\forall i \in [k], s.t. P(x_i) - P(x_{i+1}) \leq 0], \\
    C_{SS}(x,k,S) &= \mathbf{1}[\forall i \in [k], s.t. S(x_{i+1}, x_1) - S(x_i, x_1) \leq 0], \\
    C_{SD}(x,k) &= \mathbf{1}[|x_t : \forall i \in [k]| > 2], \\
    SSR &=  \frac{1}{|X|} \sum_{x \in X} \mathbf{1}[C_{SP}(x,k,P) \wedge C_{SS}(x,k,S) \wedge C_{SD}(x,k)]
\end{aligned}
\end{equation}

where $\{x_i\}_{i=1}^k$ is one $k$-step manipulation trajectory, $X$ contains $n$ manipulation trajectories, $C$ is the constraint, $P$ is the property evaluation function, $S$ is the structure similarity function (Taminoto similarity over Morgan fingerprints). The $C_{SP}$ constraints that the property of molecules must monotonically increase. The $C_{SS}$ constraints that the structure is similar in regard to the starting molecule must monotonically decrease. The $C_{SD}$ constraint that the molecules must at least change twice during the manipulation. SSR calculates the percentage of trajectories that satisfy all success constraints. 

The relaxed success rate relaxes some constraints by adding a tolerance interval. It is formulated as follows: 

\begin{equation}
\begin{aligned}
    C_{SP}(x,k,P) &= \mathbf{1}[\forall i \in [k], s.t. P(x_t) - P(x_{t+1}) \leq \epsilon], \\
    C_{SS}(x,k,S) &= \mathbf{1}[\forall i \in [k], s.t. S(x_{t+1}, x_1) - S(x_t, x_1) \leq \gamma], \\
    C_{SD}(x,k) &= \mathbf{1}[|x_t : \forall i \in [k]| > 2], \\
    SSR &=  \frac{1}{|X|} \sum_{x \in X} \mathbf{1}[C_{SP}(x,k,P) \wedge C_{SS}(x,k,S) \wedge C_{SD}(x,k)]
\end{aligned}
\end{equation}

The relaxed success rate does not require a monotonic increase of molecular property but sets a tolerance threshold $\epsilon$. This tolerance threshold $\epsilon$ is defined as $5\%$ of the range of property in the training dataset. It also does not require a monotonically decrease of structure similarity with a tolerance threshold $\gamma$ of 0.1. 

\subsection{Additional Baselines}
\label{sec:additional_baselines}

\paragraph{Evolutionary Algorithm.} 
We present the Evolutionary Algorithm-based~(EA) approach to optimize molecules in the latent space as an additional baseline for the experiment. The pseudocode is provided as \Cref{alg:ea}.
\Cref{tab:additional_baseline_results} shows the performance of the evolutionary algorithm-based approach on unconstrained optimization.
For a fair comparison, all methods in \Cref{tab:additional_baseline_results} have the same number of Oracle calls. The result shows that our methods outperform all EA approaches.

\begin{algorithm}
\caption{Evolutionary Algorithm Augmented Optimization}\label{alg:ea}
\begin{algorithmic}[1]
\STATE \textbf{Input:} $n$ samples select $k$ per iteration, pre-trained ChemSpace/Gradient direction $l$, step size $\alpha$, pre-trained surrogate model $h$, decoder $g_\psi$
\STATE Randomly sample $n$ latent vectors $\{\vz_i^0\}_{i=1}^n$ in latent space
\FOR{each iteration $t = 0$ to $T$}
    \STATE Evaluate sampled latent vector scores: $s_i^t = h(\mathbf{z}_i^t)$ for $i = 1, \ldots, n$
    \STATE Select top-$k$ scored latent vectors: $\{\vz_\text{top}^t\}_{j=1}^k$
    \FOR{each selected vector $j = 0$ to $k$}
        \STATE Update $\vz_{\text{top}, j}^{t+1} = \mathbf{z}_{\text{top}, j}^t + \alpha \cdot l + \boldsymbol{\epsilon}$, where $\boldsymbol{\epsilon} \sim \mathcal{N}(0, \mathbf{I})$, $l$ the evolution direction guided by Random/ChemSpace/Gradient.
    \ENDFOR
    \STATE Randomly sample $\frac{n}{k}$ samples around each $\vz_{\text{top}, j}^{t}$ to generate $n$ new latent vectors $\{\vz_i^{t+1}\}_{i=1}^n$
\ENDFOR
\STATE Decode latent vectors: $x_i^{T} = g_\psi(\vz_i^{T})$ for $i = 1, \ldots, n$
\end{algorithmic}
\end{algorithm}

\begin{table}[htbp]
\caption{\textbf{Unconstrained plogP, QED maximization of Evolutionary Algorithm.} 
(SPV denotes supervised scenarios, UNSUP denotes unsupervised scenarios).
Random/ChemSpace/Gradient are the evolution directions of EA.
Since EA (Gradient Flow) has converged to a single molecule when optimizing pLogP, only one value is reported.
} %
\centering
\small
\begin{sc}
\begin{tabular}{l|ccc|ccc}
\toprule
Method & \multicolumn{3}{c}{plogP $\uparrow$} & \multicolumn{3}{c}{QED $\uparrow$} \\
 & 1st & 2nd & 3rd & 1st & 2nd & 3rd \\
\midrule
EA (Random) & 2.29 & 1.64 & 1.52 & 0.836 & 0.801 & 0.794\\
EA (ChemSpace) & 3.79 & 3.79 & 3.79 & 0.933 & 0.931 & 0.931\\
EA (Gradient Flow) & 3.53 & / & / & 0.930 & 0.929 & 0.929

\\
\midrule
Wave (spv)& 4.76& 3.78& 3.71& \textbf{0.947}& 0.934& 0.932\\
 Wave (unsup)& \textbf{5.30}& \textbf{5.22}& \textbf{5.14}& 0.905& 0.902& 0.978\\
HJ (spv)& 4.39& 3.70& 3.48& 0.946& 0.941& 0.940\\
 HJ (unsup)& 4.26& 4.10& 4.07& 0.930& 0.928& 0.927\\
LD& 4.74& 3.61& 3.55& \textbf{0.947}& \textbf{0.947}& \textbf{0.942}\\
\bottomrule
\end{tabular}
\end{sc}
\label{tab:additional_baseline_results}
\end{table}

\paragraph{Unconstrained Molecular Optimization Additional Results}
In addition to reporting the top 3 scores as presented in \Cref{tab:unconstrained_optim}, we computed the mean and standard deviation for the top 100 molecules after unconstrained optimization in \Cref{tab:unconstrained_optim_mse}.
The table shows that our methods have overall the best optimization performance.
In addition, HJ exhibits better performance on mean and standard deviation than on top 3, showing that minimizing the kinetic energy is efficient in pushing the distribution to desired properties.

\begin{table}[htbp]
\caption{\textbf{MSE of Unconstrained plogP, QED maximization, and docking score minimization.}
(SPV denotes supervised scenarios, UNSUP denotes unsupervised scenarios).
Each entry in the table follows the format mean ± std (median).
Boldface highlights the highest-performing generation for each property within each rank.
} %
\centering
\small
\resizebox{\linewidth}{!}{ %
\begin{sc}
\begin{tabular}{@{}c|cccc@{}}
\toprule
Method& plogP $\uparrow$ & QED $\uparrow$& ESR1 Docking  $\downarrow$ & ACAA1 Docking $\downarrow$ \\ \midrule
Random & 2.345 ± 0.386 (2.259)& 0.903 ± 0.014 (0.902)& -9.127 ± 0.360 (-9.015)& -8.454 ± 0.316 (-8.390)\\
\midrule
Gradient Flow & 2.664 ± 0.382 (2.537)& 0.910 ± 0.012 (0.908)& -9.452 ± 0.338 (-9.365)& -8.735 ± 0.337 (-8.650)\\
ChemSpace & 2.580 ± 0.406 (2.446)& 0.907 ± 0.014 (0.906)& -9.523 ± 0.409 (-9.395)& -8.749 ± 0.356 (-8.640)\\
\midrule
Wave (spv) & 2.536 ± 0.439 (2.388)& 0.903 ± 0.015 (0.898)& \textbf{-9.630 ± 0.399 (-9.525)}& -8.764 ± 0.344 (-8.650)\\
Wave (unsup) & 1.736 ± 0.401 (1.610)& 0.845 ± 0.014 (0.840)& -9.074 ± 0.329 (-9.000)& \textbf{-8.813 ± 0.265 (-8.745)}\\
HJ (spv) & 2.482 ± 0.397 (2.382)& 0.899 ± 0.017 (0.894)& -9.544 ± 0.322 (-9.460)& -8.792 ± 0.332 (-8.675)\\
HJ (unsup) & \textbf{3.405 ± 0.254 (3.377)}& \textbf{0.911 ± 0.009 (0.909)}& -9.132 ± 0.321 (-9.090)& -8.668 ± 0.243 (-8.630)\\
 LD & 2.463 ± 0.388 (2.399)& 0.905 ± 0.014 (0.903)& -9.400 ± 0.360 (-9.300)&-8.709 ± 0.372 (-8.585)\\
\bottomrule
\end{tabular}
\end{sc}
} %
\label{tab:unconstrained_optim_mse}
\end{table}

\subsection{More Experiment Results}
\label{sec:more_exp}

We conduct more experiments to analyze the performance of the proposed methods systematically. They are referred to and discussed in the main paper.

\paragraph{Pearson correlation score for unsupervised diversity guidance with disentanglement regularization.}
\Cref{tab:unsupervised_corr,tab:unsupervised_corr_hj} present the Pearson correlation scores for the trained energy networks of wave equations and Hamilton-Jacobi equations with disentanglement regularization, respectively. For properties other than synthetic accessibility (SA), we select the network with the highest correlation score to maximize these properties.
Conversely, for SA, the network with the lowest correlation score (most negative score) is chosen.
\begin{table}[htbp]
\caption{ 
\textbf{Pearson Correlation of trained Wave PDE Energy Network.} The average Pearson correlation between the sequence of real properties and sequence of time steps along the manipulation trajectory following a learned potential function $\phi^k(t, z)$ using wave equations.
} %
\centering
\small
\begin{sc}
\begin{tabular}{@{}c|cccccc@{}}
\toprule
 & plogP & SA & QED & DRD2 & JNK3 & JSK3B \\ \midrule
0 & 0.019& 0.003& 0.016& 0.029& 0.015& \textbf{0.051}\\
1 & \textbf{0.160}& \textbf{-0.451}& \textbf{0.275}& -0.074& -0.153& -0.272\\
2 & 0.035& -0.003& 0.011& 0.002& -0.006& 0.017\\
3 & 0.072& -0.096& 0.065& 0.011& -0.017& -0.028\\
4 & 0.042& 0.003& 0.039& -0.010& \textbf{0.025}& 0.018\\
5 & -0.036& -0.022& 0.150& 0.009& -0.017& 0.008\\
6 & 0.032& -0.045& 0.006& 0.002& -0.011& 0.002\\
7 & 0.023& -0.023& 0.054& -0.002& -0.013& -0.017\\
8 & 0.075& -0.085& 0.064& -0.040& -0.007& -0.054\\
9 & 0.013& 0.020& -0.011& \textbf{0.031}& 0.005& 0.014\\ \midrule
index & 1& 1& 1& 9& 4& 0\\ \bottomrule
\end{tabular}
\end{sc}
\label{tab:unsupervised_corr}
\end{table}

\begin{table}[htbp]
\caption{
\textbf{Pearson Correlation of trained Hamilton-Jacobian PDE Energy Network.} The average Pearson correlation between the sequence of real properties and sequence of time steps along the manipulation trajectory following a learned potential function $\phi^k(t,z)$ using Hamilton-Jacobi equations.
} %
\centering
\small
\begin{sc}
\begin{tabular}{@{}c|cccccc@{}}
\toprule
 & plogP & SA & QED & DRD2 & JNK3 & GSK3B \\ \midrule
0 & \textbf{0.345}& \textbf{-0.453}& 0.141& -0.210& -0.127& -0.350\\
1 & 0.257& -0.289& 0.057& -0.154& -0.121& -0.276\\
2 & 0.203& -0.284& 0.050& \textbf{0.020}& -0.044& -0.164\\
3 & -0.029& 0.041& 0.034& -0.001& \textbf{0.001}& 0.002\\
4 & 0.304& -0.343& 0.066& -0.225& -0.179& -0.336\\
5 & 0.008& -0.036& 0.029& -0.009& -0.021& 0.015\\
6 & 0.316& -0.370& 0.173& -0.208& -0.163& -0.309\\
7 & 0.305& -0.429& 0.046& -0.291& -0.191& -0.352\\
8 & 0.003& -0.011& 0.016& -0.021& -0.035& \textbf{0.026}\\
9 & 0.311& -0.386& \textbf{0.209}& -0.222& -0.209& -0.342\\ \midrule
index & 0& 0& 9& 2& 3& 8\\ \bottomrule
\end{tabular}
\end{sc}
\label{tab:unsupervised_corr_hj}
\end{table}

\begin{figure}[htbp]
\begin{center}
\includegraphics[width=\linewidth]{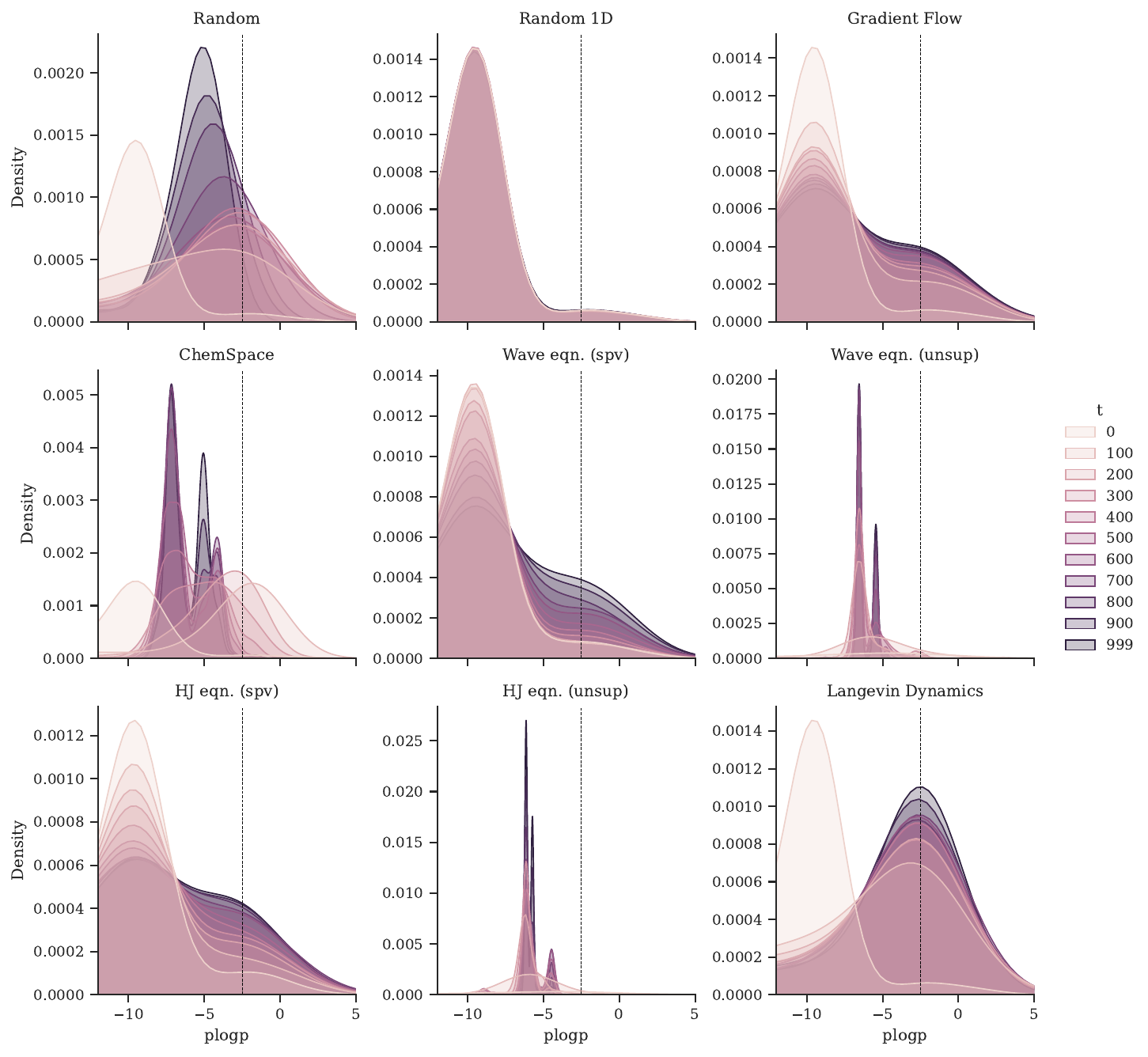}
\end{center}
\caption{
\textbf{Distribution shift for plogP optimization}
}
\label{fig:plogp_kde}
\end{figure}

\begin{figure*}[htbp]
\begin{center}
\includegraphics[width=\linewidth]{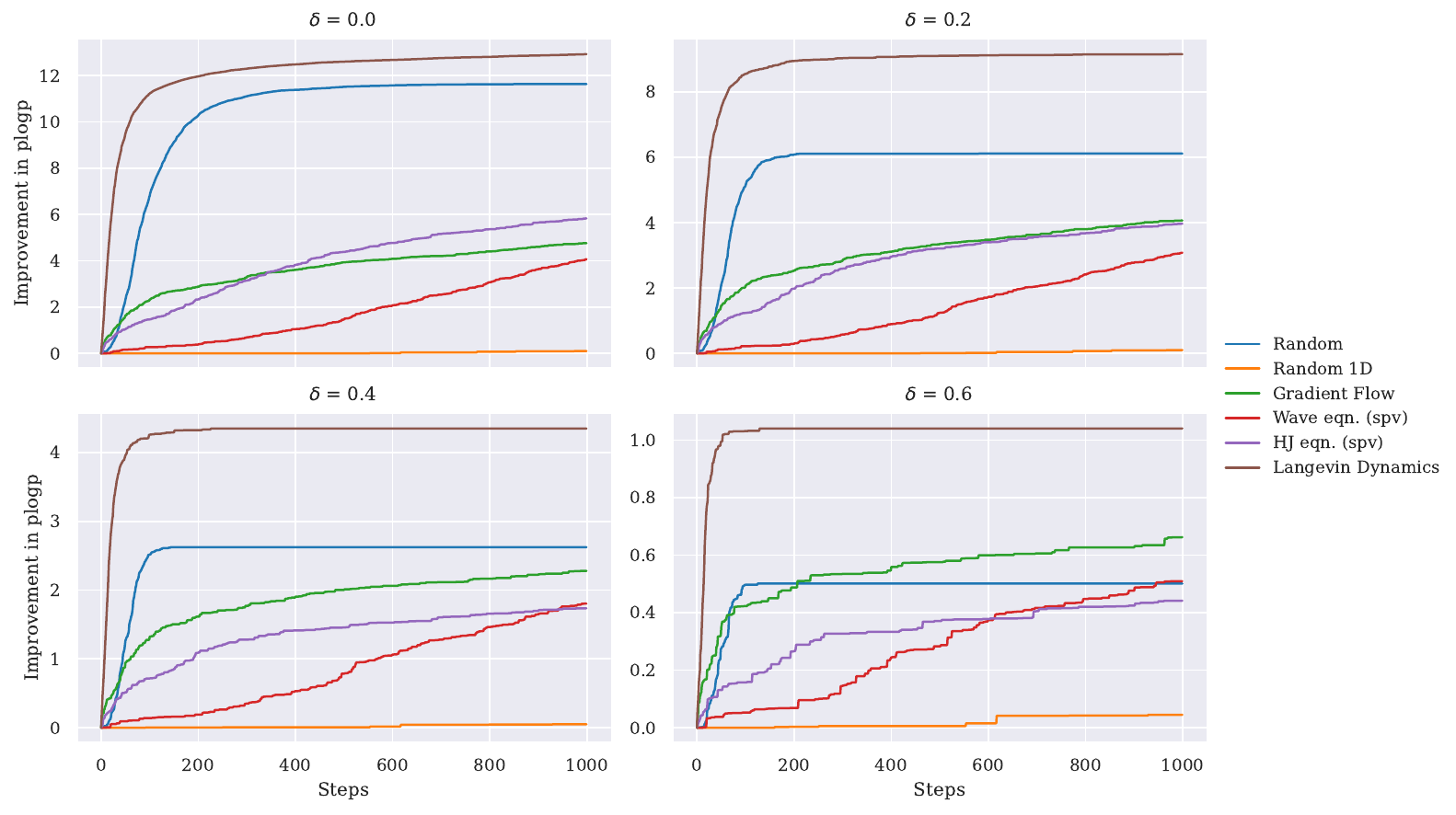}
\end{center}
\caption{
\textbf{Optimization Convergence} Langevin Dynamics shows faster convergence and achieves greater improvement in plogP.
}
\label{fig:optim_plogp_conv}
\end{figure*}

\paragraph{Distribution shift and convergence for plogP optimization.}
\Cref{fig:plogp_kde} illustrates the distribution shift in plogP optimization, complementing the analysis in \Cref{fig:plogp_spv_kde}.
Similar to the findings discussed in \Cref{sec:mol_optim}, both unsupervised methods encounter out-of-distribution (OOD) issues after 400 steps, consistent with those observed with ChemSpace.
The Random 1D method does not achieve the expected distribution shift, as it manipulates only one dimension of the latent vector.
\Cref{fig:optim_plogp_conv} depicts the convergence trends of each method's improvements.
Consistent with the predictions in \Cref{prop:global_converge}, Langevin Dynamics demonstrates the fastest and most effective convergence among all methods.

\begin{figure}[htbp]
\begin{center}
\includegraphics[width=\linewidth]{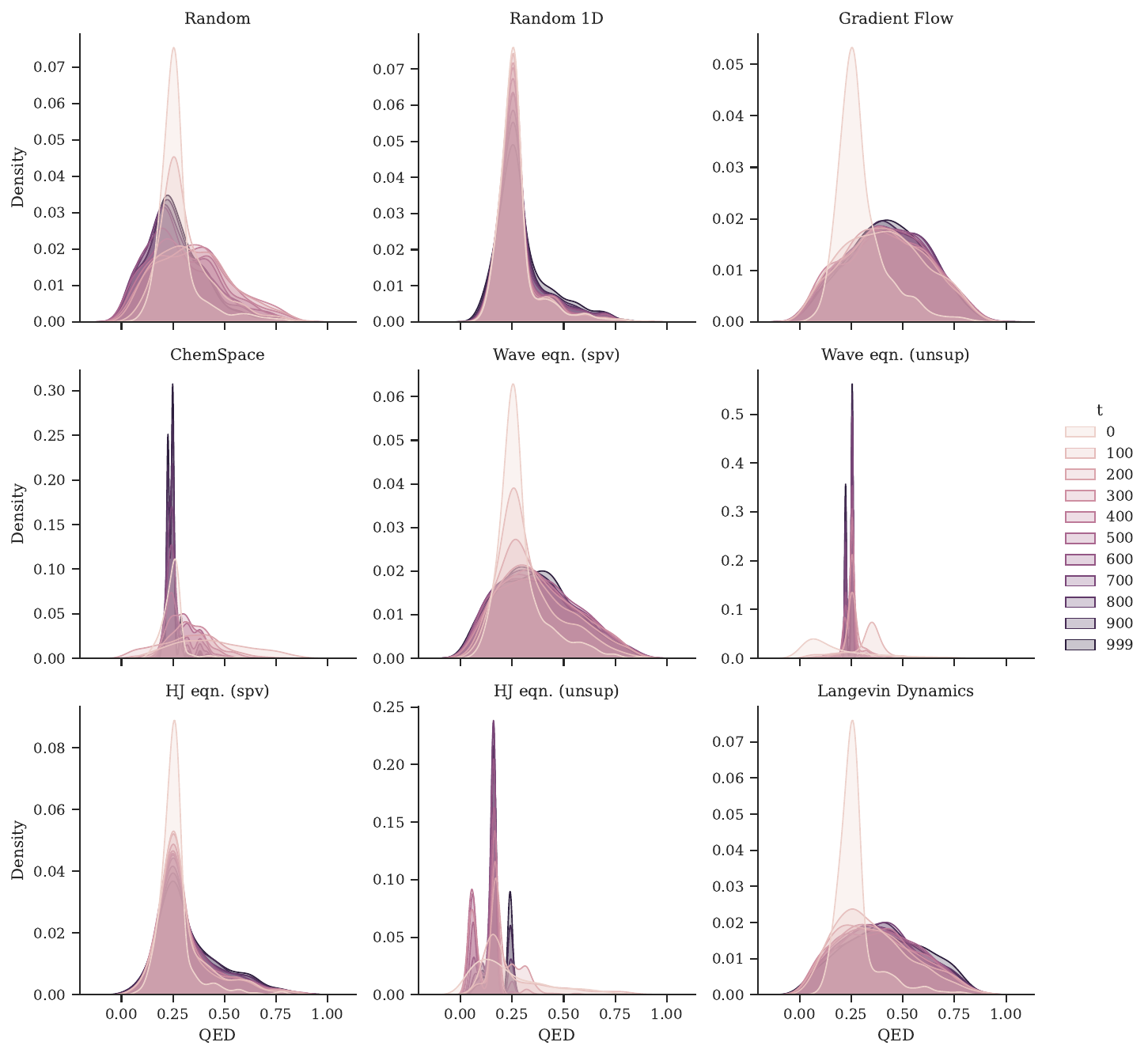}
\end{center}
\caption{
\textbf{Distribution shift for QED optimization}
}
\label{fig:qed_kde}
\end{figure}

\paragraph{Similarity-constrained QED optimization and distribution shift.}
To further explore optimization tasks on additional properties, we also attempted to enhance the QED of molecules.
\Cref{tab:optim_qed} presents the results of efforts to maximize the QED of 800 molecules from the ZIC250K dataset that initially had the lowest QED scores.
Despite encountering OoD issues with ChemSpace and two unsupervised methods, as illustrated in \Cref{fig:qed_kde}, the performance of Langevin Dynamics surpasses other methods across various similarity levels, which is consistent with the result from \Cref{tab:optim_plogp}.
\begin{table}[htbp]
\caption{
\textbf{Similarity-constrained QED maximization.}
For each method with minimum similarity constraint $\delta$, the results in reported in format mean ± standard derivation (success rate \%) of absolute improvement, where the mean and standard derivation are calculated among molecules that satisfy the similarity constraint.
} %
\centering
\resizebox{0.8\linewidth}{!}{ %
\begin{tabular}{@{}c|cccc@{}}
\toprule
Method& $\delta = 0$ & $\delta = 0.2$ & $\delta = 0.4$ & $\delta = 0.6$ \\ \midrule
Random & 0.36 ± 0.15 (98.0)& 0.19 ± 0.14 (78.4)& 0.11 ± 0.11 (54.4)& 0.08 ± 0.08 (29.9)\\
Random 1D& 0.13 ± 0.11 (40.1)& 0.12 ± 0.10 (38.1)& 0.09 ± 0.07 (29.2)& 0.07 ± 0.06 (18.0)\\
\midrule
Gradient Flow & 0.48 ± 0.13 (99.2)& 0.25 ± 0.16 (84.9)& 0.13 ± 0.12 (53.8)& 0.10 ± 0.10 (24.9)\\
ChemSpace & 0.47 ± 0.13 (99.8)& 0.29 ± 0.18 (90.1)& 0.18 ± 0.15 (62.9)& 0.12 ± 0.12 (33.5)\\
\midrule
Wave (spv) & 0.38 ± 0.16 (97.9)& 0.23 ± 0.15 (84.9)& 0.13 ± 0.11 (62.6)& 0.08 ± 0.08 (35.0)\\
Wave (unsup) & \textbf{0.54 ± 0.18 (99.1)}& 0.09 ± 0.10 (51.0)& 0.05 ± 0.06 (27.4)& 0.03 ± 0.04 (14.0)\\
HJ (spv) & 0.21 ± 0.16 (74.2)& 0.17 ± 0.14 (69.4)& 0.12 ± 0.11 (57.0)& 0.08 ± 0.09 (35.0)\\
HJ (unsup) & 0.52 ± 0.19 (98.5)& 0.11 ± 0.11 (60.4)& 0.05 ± 0.05 (28.6)& 0.03 ± 0.03 (13.4)\\
 LD & 0.53 ± 0.12 (99.8)& \textbf{0.31 ± 0.17 (96.2)}& \textbf{0.16 ± 0.12 (77.8)}&\textbf{0.10 ± 0.09 (49.5)}\\
\bottomrule
\end{tabular}
} %
\label{tab:optim_qed}
\end{table}

\section{Latent Space Visualization and Analysis}
\label{sec:latent_viz}

\begin{figure*}[htbp]
\begin{center}
\includegraphics[width=0.5\linewidth]{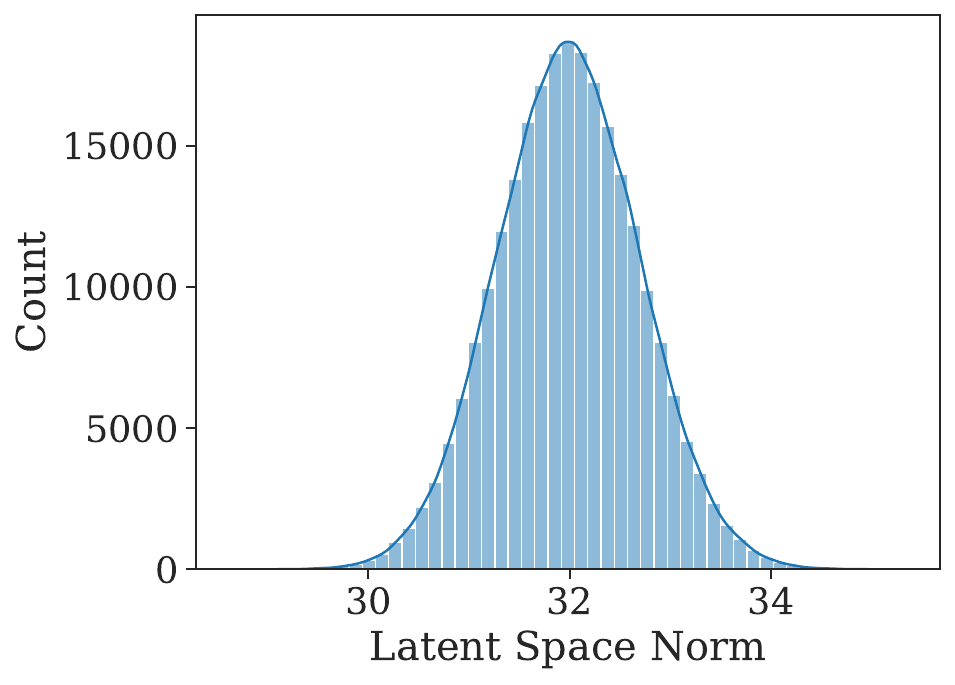}
\end{center}
\caption{\textbf{Latent Vector Norm. } Distribution of the norm of the latent vectors projected from the training dataset onto the learned latent space.}
\label{fig:latent_norm}
\end{figure*}

\begin{figure*}[htbp]
\begin{center}
\includegraphics[width=\linewidth]{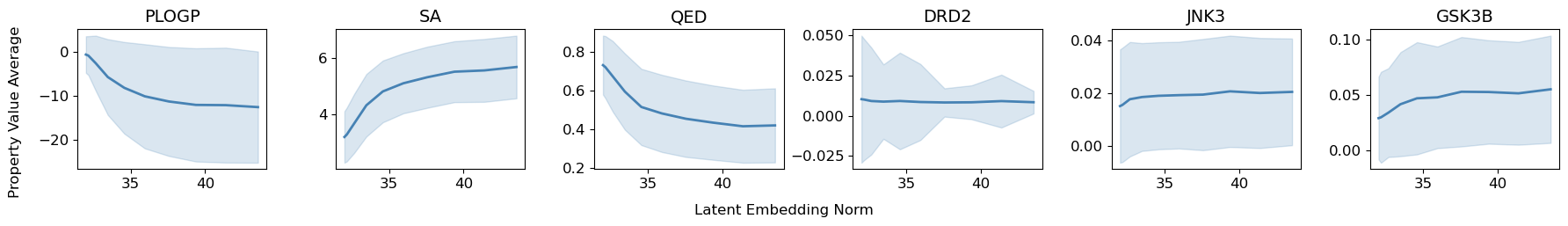}
\end{center}
\caption{\textbf{Embedding Norm against Property Value of each path.} Norm and property value of molecules along the direction of latent traversal with a random direction. The middle curve shows the mean property value and latent embedding norm for all paths. The shaded area is the standard deviation of property value.}
\label{fig:prop_vs_latent_embedding_by_step}
\end{figure*}

\begin{figure*}[htbp]
\begin{center}
\includegraphics[width=\linewidth]{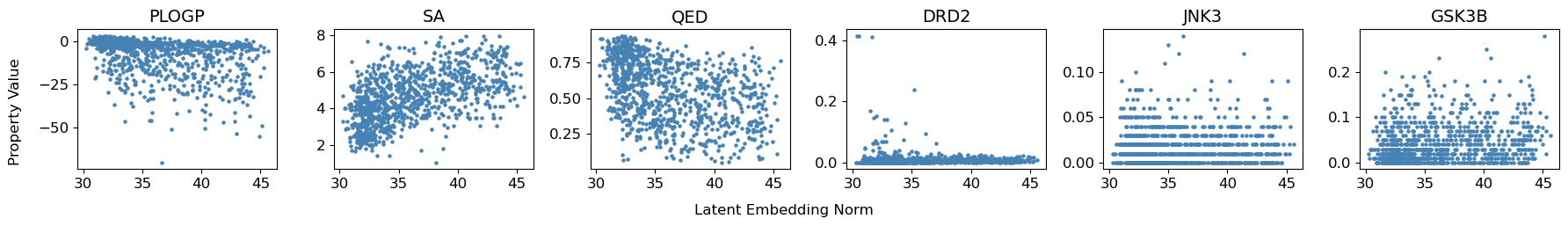}
\end{center}
\caption{\textbf{Embedding Norm against Property Value of each Molecule.}  Scatter plot of norm and property value of individual molecules in the training set encoded in the latent space.}
\label{fig:prop_vs_latent_embedding_scatter}
\end{figure*}

As observed in experiments such that random directions perform surprisingly well on molecule manipulation and optimization tasks, we look into the learned latent space to understand its structure. As the prior of a VAE is an isotropic Gaussian distribution, we first verify if the learned variational poster also follows a Gaussian distribution and we find that it does learn so from the evidence shown in \Cref{fig:latent_norm}, where the norm of the molecule projected to the latent space concentrate around 32 which is around $\sqrt{d}$ such that the latent dimension $d$ is 1024. We also visualize in \Cref{fig:prop_vs_latent_embedding_by_step} how the properties of the molecules in the training dataset are related to their latent vector norms. Surprisingly, we find a strong correlation between almost all molecular properties and their latent norms. Combining these two evidences, it is not surprising that a random latent vector taking a random direction will change the molecular property smoothly and monotonically. In addition, we further plot when we traverse along a random direction in the latent space, how the change of the norm may correspond to the change of a certain property. Among them, we find that SA is particularly in strong positive correlation with the traversal in \Cref{fig:prop_vs_latent_embedding_scatter}. Though the emergence of the structure in the latent space is interesting and suggests that better algorithms can be developed to exploit the structure, we leave this to future work.

\begin{figure*}[htbp]
\begin{center}
\includegraphics[width=\linewidth]{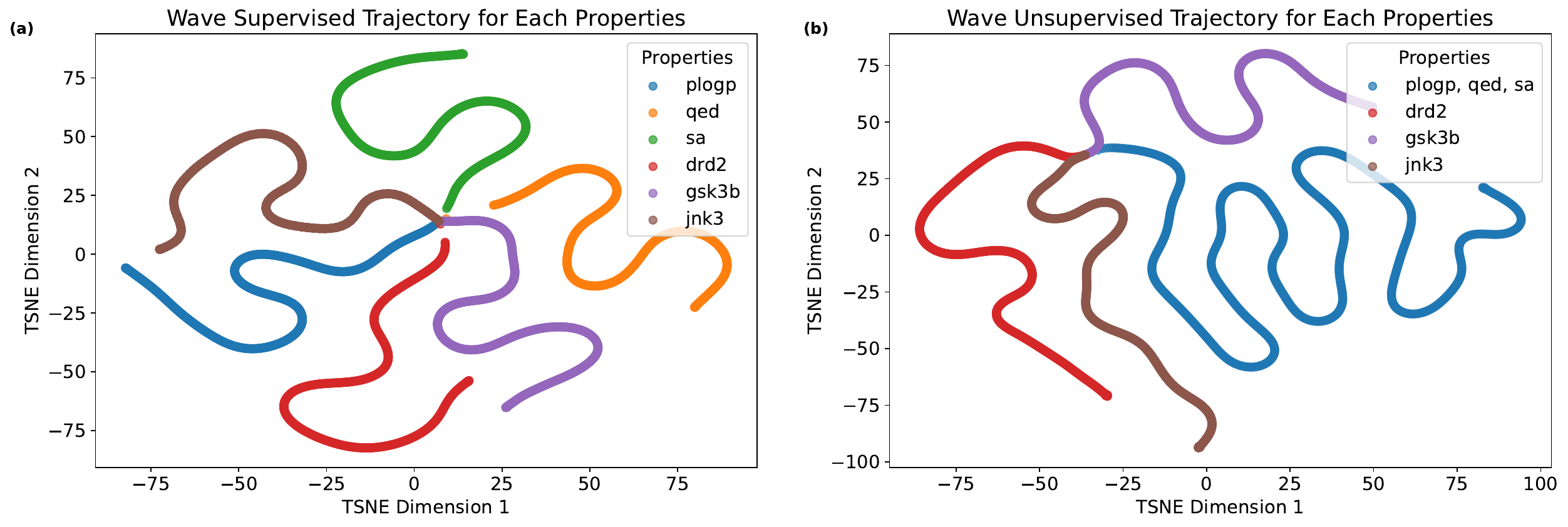}
\end{center}
\caption[Two numerical solutions]{\textbf{t-SNE Visualization of Optimization Trajectories. } Optimization Trajectories following supervised and unsupervised wave flow visualized using t-SNE.}
\label{fig:wave_sup_unsup_tsne}
\end{figure*}

Additionally, we visualize the traversal trajectory for each different property using both supervised and unsupervised wave flow in \Cref{fig:wave_sup_unsup_tsne}.
The plot shows that almost all trajectories grow towards a unique direction in the t-SNE plot, which implies the disentanglement of learned directions and, thus, molecular properties.
In addition, the figures display sinusoidal wave-shape trajectories, indicating the flow follows wave-like dynamics.
In the unsupervised t-SNE plot, the trajectories of some properties overlap, such as plogP and SA.
It is because some properties correlate with the same disentangled direction, so their traversal follows the same direction, thus the same trajectories.

\section{Qualitative Evaluations}
\label{sec:more_viz}

\begin{figure*}[htbp]
\begin{center}
\includegraphics[width=\linewidth]{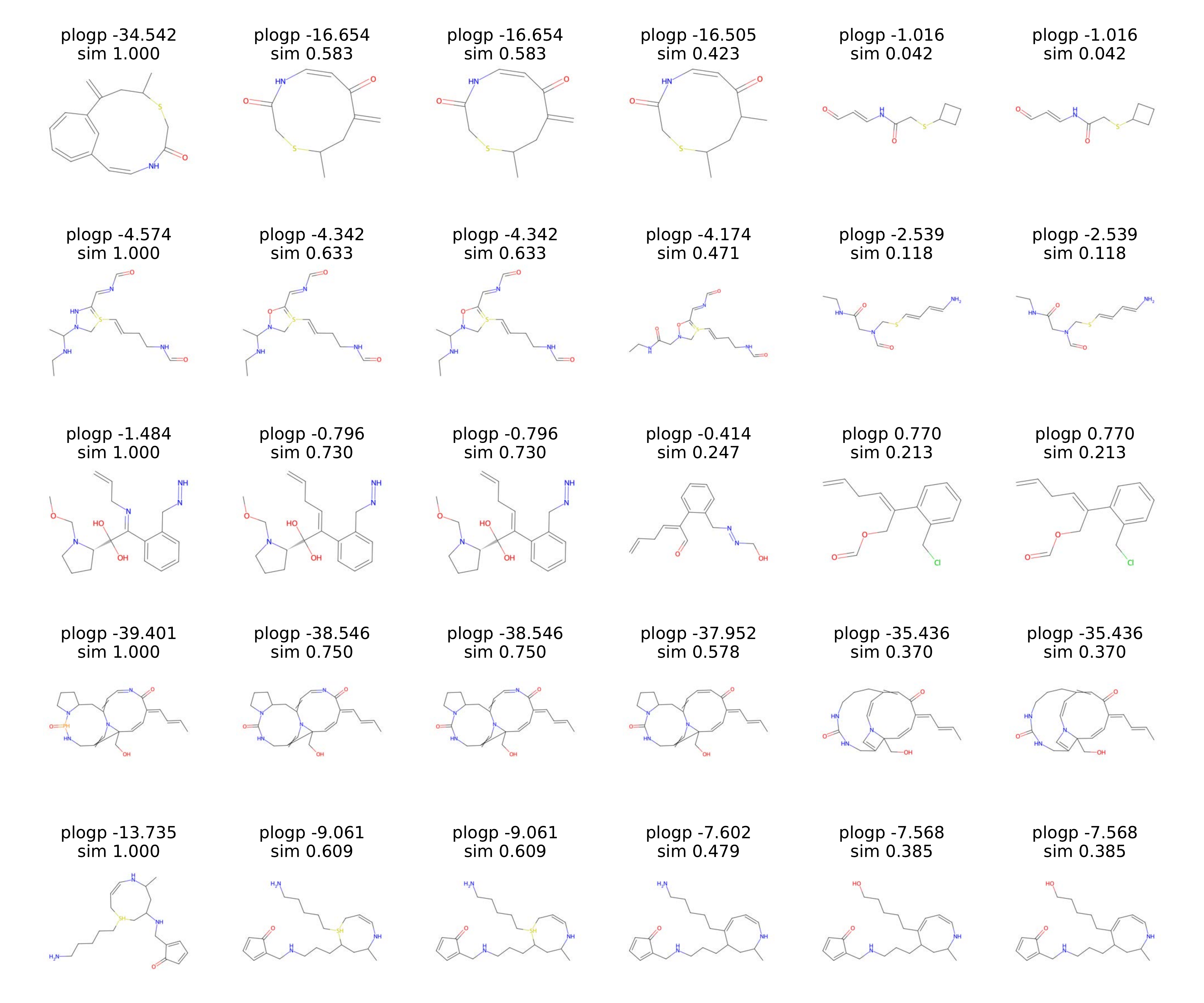}
\end{center}
\vspace{-0.7cm}
\caption{\textbf{Molecule Manipulation Trajectory}
Molecule manipulation by chemspace on plogP.}
\label{fig:manipulation_plogp_chemspace}
\end{figure*}

\begin{figure*}[htbp]
\begin{center}
\includegraphics[width=\linewidth]{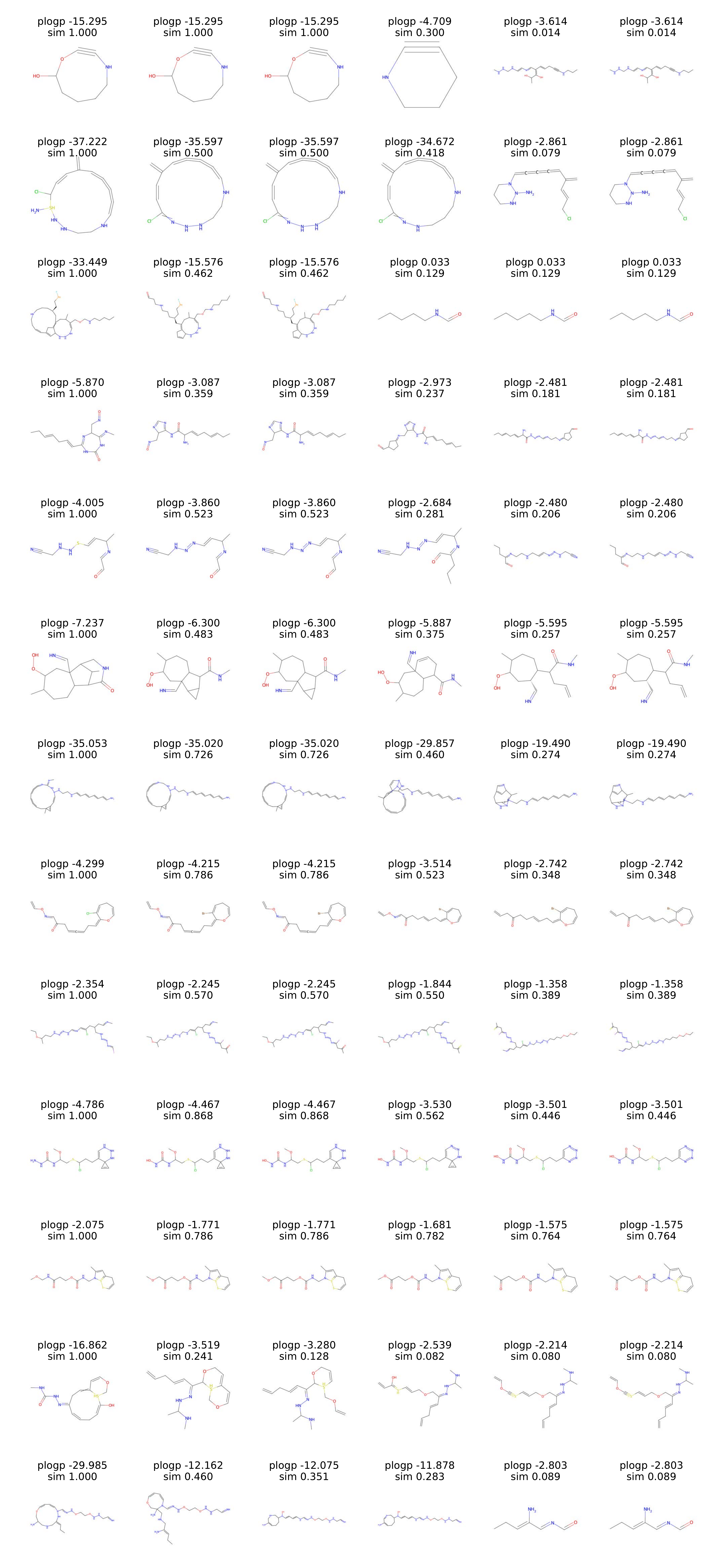}
\end{center}
\vspace{-0.7cm}
\caption{\textbf{Molecule Manipulation Trajectory}
Molecule manipulation by gradient flow on plogP.}
\label{fig:manipulation_plogp_limo}
\end{figure*}

\begin{figure*}[htbp]
\begin{center}
\includegraphics[width=\linewidth]{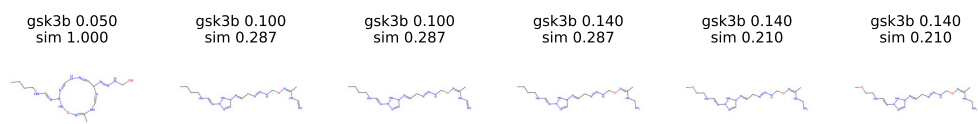}
\end{center}
\vspace{-0.7cm}
\caption{\textbf{Molecule Manipulation Trajectory}
Molecule manipulation by supervised wave flow on GSK3B.}
\label{fig:manipulation_gsk3b_wavepde}
\end{figure*}

\begin{figure*}[htbp]
\begin{center}
\includegraphics[width=\linewidth]{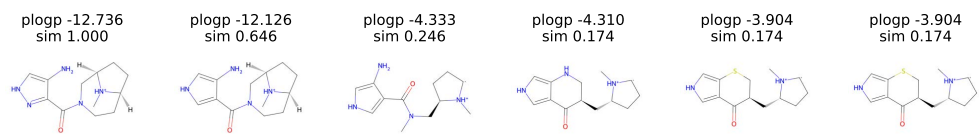}
\end{center}
\vspace{-0.7cm}
\caption{\textbf{Molecule Optimization Trajectory}
Molecule optimization by random direction on plogP.}
\label{fig:optimization_plogp_random}
\end{figure*}

\begin{figure*}[htbp]
\begin{center}
\includegraphics[width=\linewidth]{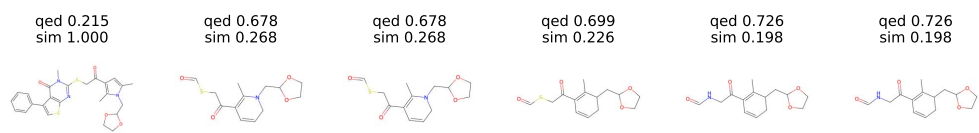}
\end{center}
\vspace{-0.7cm}
\caption{\textbf{Molecule Optimization Trajectory}
Molecule optimization by supervised wave flow on QED.}
\label{fig:optimization_qed_wavepde}
\end{figure*}

\begin{figure*}[htbp]
\begin{center}
\includegraphics[width=\linewidth]{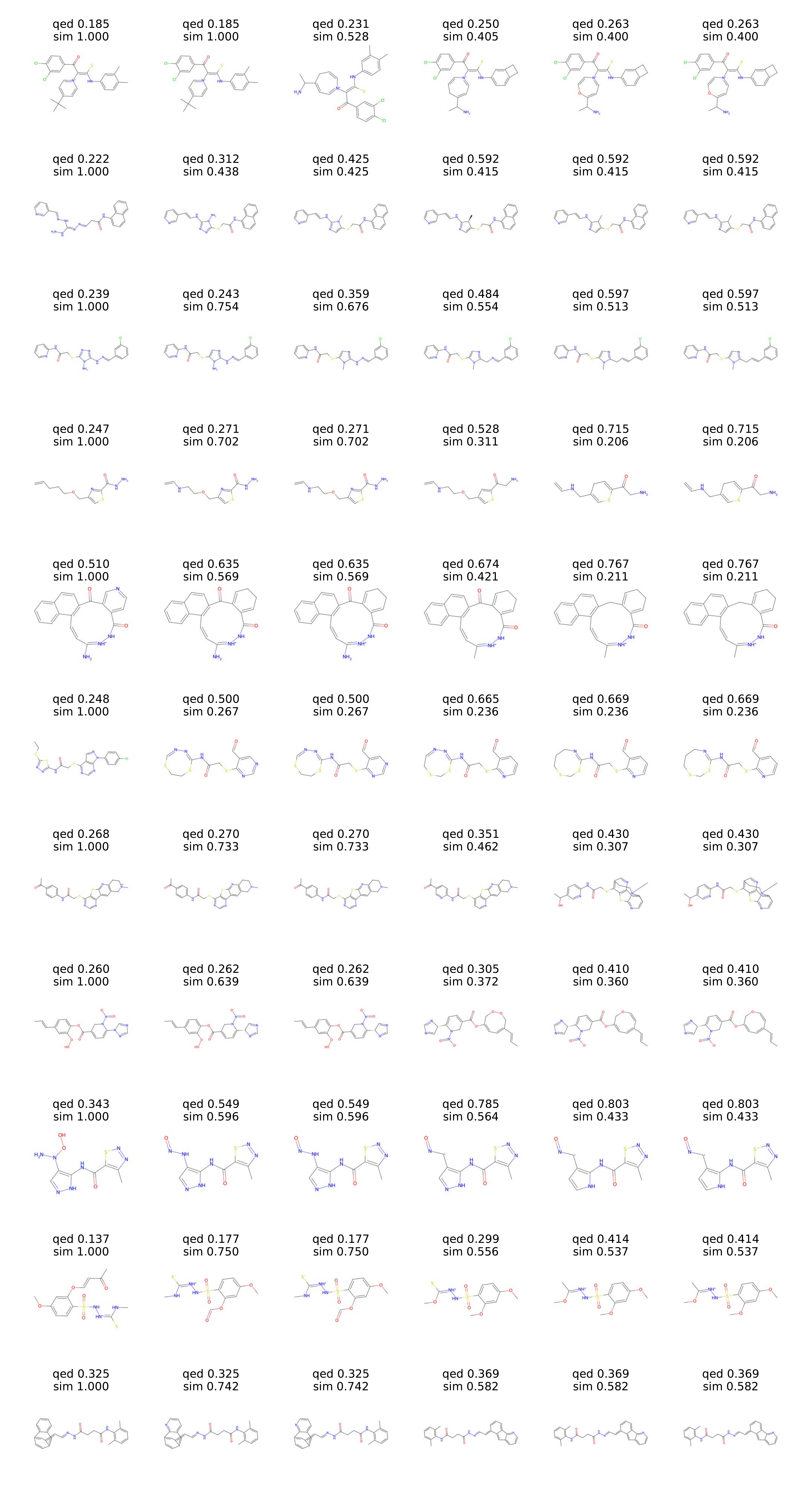}
\end{center}
\vspace{-0.7cm}
\caption{\textbf{Molecule Optimization Trajectory}
Molecule optimization by supervised Hamilton-Jacobi flow on QED.}
\label{fig:optimization_qed_hj}
\end{figure*}

In addition to quantitative evaluations, we demonstrate some qualitative evaluations in this section. 
We showcase three manipulation trajectories in \cref{fig:manipulation_plogp_chemspace,fig:manipulation_plogp_limo,fig:manipulation_gsk3b_wavepde}. Each of these paths is a 6-step manipulation for different molecular properties using different flows. Specifically, we can see that the supervised wave flow and gradient flow improves the molecular property by the conversion of large heterocyclic rings for better synthesizability. We also show three optimization trajectories in \cref{fig:optimization_plogp_random,fig:optimization_qed_wavepde,fig:optimization_qed_hj}. From left to right, each molecule is a snapshot selected from a trajectory of 1000-step optimization. It is notable that the supervised Hamilton-Jacobi flow optimizes the property by reducing the number of nitrogen atoms. This leads to a more chemically stable molecule, whereas the original molecule, with a large number of nitrogen atoms, is unstable and potentially explosive. The supervised wave flow optimizes the molecular property by simplifying the poly-cyclic molecule which enhances its synthesizability and overall stability.

\end{document}